\documentclass{article}

\PassOptionsToPackage{numbers, compress}{natbib}

\usepackage[preprint]{neurips_2019}

\usepackage[utf8]{inputenc} %
\usepackage[T1]{fontenc}    %
\usepackage{hyperref}       %
\usepackage{url}            %
\usepackage{booktabs}       %
\usepackage{amsfonts}       %
\usepackage{nicefrac}       %
\usepackage{microtype}      %
\usepackage{graphicx}
\usepackage{subfigure}
\usepackage{color}
\usepackage{multirow}
\usepackage{wrapfig}
\usepackage{amsmath}
\usepackage{soul}
\usepackage{verbatim}
\usepackage[bottom]{footmisc}

\usepackage{xcolor}
\definecolor{mydarkblue}{rgb}{0,0.08,0.45}
\hypersetup{ %
pdftitle={},
pdfauthor={},
pdfsubject={},
pdfkeywords={},
pdfborder=0 0 0,
pdfpagemode=UseNone,
colorlinks=true,
linkcolor=mydarkblue,
citecolor=mydarkblue,
filecolor=mydarkblue,
urlcolor=mydarkblue,
pdfview=FitH}

\newcommand{\aug}[1]{\texttt{#1}}

\DeclareMathOperator*{\expectation}{\mathop{\mathbb{E}}}

\newcommand{\linebreakcell}[2][c]{%
  \begin{tabular}[#1]{@{}c@{}}#2\end{tabular}}
  
\newcommand{\leftlinebreakcell}[2][c]{%
  \begin{tabular}[#1]{@{}l@{}}#2\end{tabular}}

\title{Improving Robustness Without Sacrificing Accuracy with Patch Gaussian Augmentation}

\author{%
  Raphael Gontijo Lopes\textsuperscript{1}\thanks{Work done as a member of the Google AI Residency program \texttt{g.co/airesidency}.}, \,Dong Yin\textsuperscript{2}\thanks{Work completed during internship at Google Brain.}, \,Ben Poole\textsuperscript{1}, \,Justin Gilmer\textsuperscript{1}, \,Ekin D. Cubuk\textsuperscript{1} \\
  \textsuperscript{1}Google Brain, \,\textsuperscript{2}UC Berkeley\\
  \texttt{\{iraphael,pooleb,gilmer,cubuk\}@google.com, dongyin@berkeley.edu}%
}

\begin{document}

\maketitle

\begin{abstract}

Deploying machine learning systems in the real world requires both high accuracy on clean data and robustness to naturally occurring corruptions. While architectural advances 
have led to improved accuracy, building robust models remains challenging.
Prior work has argued that there is an inherent trade-off between 
robustness and accuracy, which is exemplified by standard data augment techniques such as \aug{Cutout}, which improves clean accuracy but not robustness, and additive \aug{Gaussian} noise, which improves robustness but hurts accuracy.
To overcome this trade-off, we introduce \aug{Patch Gaussian}, a simple augmentation scheme that adds noise to randomly selected patches in an input image.
Models trained with \aug{Patch Gaussian} achieve state of the art on the CIFAR-10 and ImageNet Common Corruptions benchmarks while also improving accuracy on clean data.
We find that this augmentation leads to reduced sensitivity to high frequency noise (similar to \aug{Gaussian}) while retaining the ability to take advantage of relevant high frequency information in the image (similar to \aug{Cutout}).
Finally, we show that \aug{Patch Gaussian} can be used in conjunction with other regularization methods
and data augmentation policies such as
AutoAugment, and improves performance on the COCO object detection benchmark.
\end{abstract}
\section{Introduction}
\label{intro}

\begin{wrapfigure}{r}{0.476\textwidth}%
\vspace{-0.8in}
\begin{center}%
\includegraphics[width=0.4\textwidth]{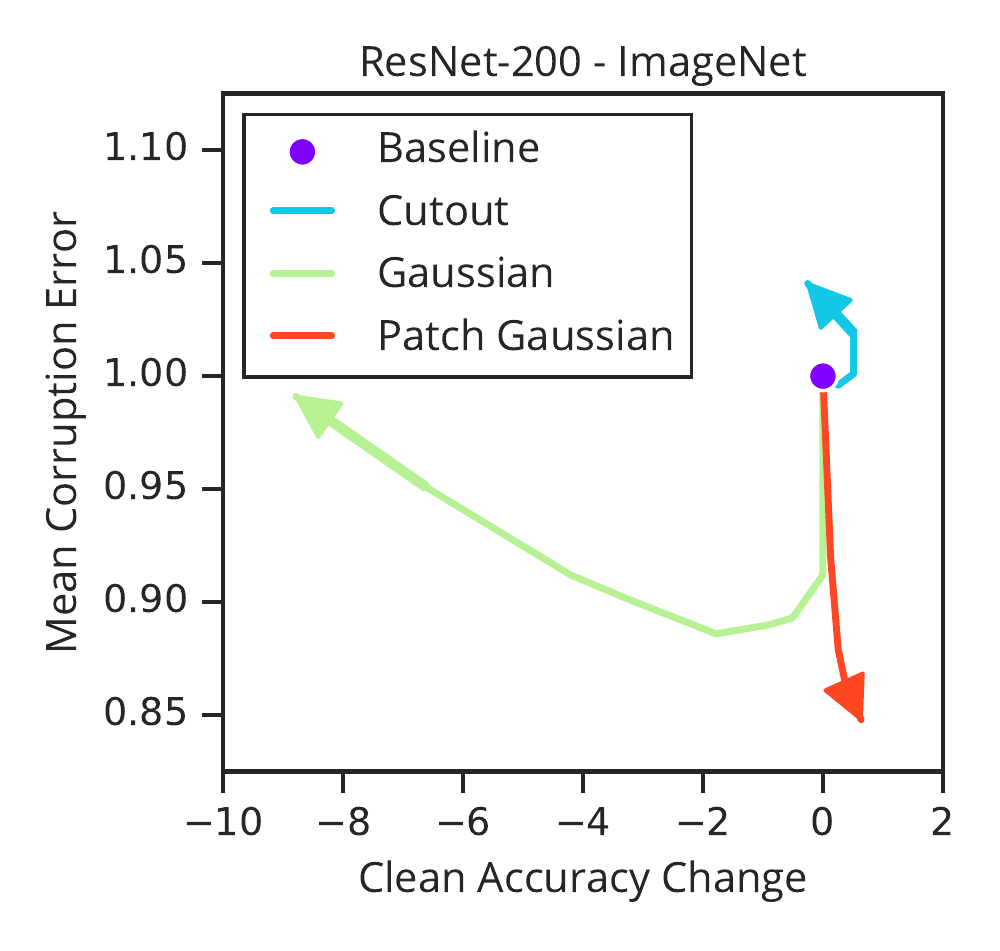}
\vspace{-0.15in}
\caption{\aug{Patch Gaussian} augmentation overcomes the accuracy/robustness trade-off observed in other augmentation strategies. Larger $\sigma$ of \aug{Patch Gaussian} improves both mean corruption error (mCE) and clean accuracy, whereas larger $\sigma$ of \aug{Gaussian} and patch size of \aug{Cutout} hurt accuracy or robustness. 
More robust and accurate models are down and to the right.
}
\label{summary}%
\end{center}
\vspace{-0.4in}
\end{wrapfigure} 

Modern deep neural networks can achieve impressive performance at classifying images in curated datasets~\cite{karpathy2011lessons, krizhevsky2012imagenet, huang2018gpipe}. Yet their performance is not robust to corruptions that typically occur in real-world settings. For example, neural networks are sensitive to small translations and changes in scale~\cite{azulay2018deep}, blurring and additive noise~\cite{dodge2017study}, small objects placed in images~\cite{rosenfeld2018elephant}, and even different images from a similar distribution of the training set~\cite{recht2019imagenet, recht2018cifar}.
For models to be useful in the real world, they need to be both accurate on a high-quality held-out set of images
, which we refer to as ``clean accuracy,''
 and robust on corrupted images, which we refer to as ``robustness.''
Most of the literature in machine learning has focused on architectural changes~\cite{simonyan2014very,szegedy2015going,he2016deep, zoph2016neural, szegedy2017inception,han2017deep,zoph2017learning,hu2017squeeze, liu2018darts} to 
improve clean accuracy
but has recently become interested in robustness as well.

Research in neural network robustness has tried to quantify the problem by establishing benchmarks that directly measure it~\cite{hendrycks2018benchmarking, gu2019using} and comparing the performance of humans and neural networks~\cite{geirhos2018generalisation}. 
Others have tried to understand robustness by highlighting systemic failure modes of current learning methods. 
For instance, networks exhibit excessive invariance to visual features~\cite{jacobsen2018excessive}, texture bias~\cite{geirhos2018imagenet}, sensitivity to worst-case (adversarial) perturbations \cite{goodfellow2014explaining}, and a propensity to rely solely on non-robust, but highly predictive features for classification~\cite{doersch2015unsupervised, ilyas2019adversarial}. Of particular relevance to our work, Ford et al \citep{ford2019adversarial} show that in order to get adversarial robustness, one needs
robustness to noise-corrupted data.

Another line of work has attempted to increase model robustness performance, either by directly projecting out superficial statistics~\cite{wang2019learning}, via architectural improvements~\cite{cubuk2017intriguing}, pre-training schemes~\cite{hendrycks2019using}, or through the use of data augmentations. Data augmentation increases the size and diversity of the training set, and provides a simple method for learning invariances that are challenging to encode architecturally \cite{cubuk2018autoaugment}.
Recent work in this area includes learning better transformations \cite{devries2017improved, zhang2017mixup, zhong2017random}, inferring combinations of transformations~\cite{cubuk2018autoaugment}, unsupervised methods~\cite{1904.12848}, theory of data augmentation \cite{dao2018kernel}, and applications for one-shot learning \cite{1904.13132}.%

Despite these advances, individual data augmentation methods that improve robustness do so at the expense of reduced clean accuracy.
Some have even claimed that there exists a fundamental trade-off between the two~\cite{tsipras2018robustness}. Because of this, many recent works focus on improving either one or the other~\cite{madry2017towards, geirhos2018imagenet}.
In this work we propose a data augmentation that overcomes this trade-off, achieving both improved robustness and clean accuracy. Our contributions are as follows:
\begin{itemize}
    \itemsep -0.0em
    \item We characterize a trade-off between robustness and accuracy among two standard data augmentations: \aug{Cutout} and \aug{Gaussian} (Section \ref{section-tradeoff}).%
    \item We devise a simple data augmentation method (which we term \aug{Patch Gaussian}) that allows us to interpolate between the two augmentations above. (Section \ref{section-patch-gauss})%
    \item We find that \aug{Patch Gaussian} allows us to overcome the observed trade-off (Figure~\ref{summary}, Section \ref{section-overcoming}), and achieves a new state of the art in the Common Corruptions benchmark \cite{hendrycks2018benchmarking} on CIFAR-C and ImageNet-C. (Section \ref{section-sota-on-cc})
    \item We demonstrate that \aug{Patch Gaussian} can be combined with other regularization strategies (Section \ref{compare-other-regs}) and data augmentation policies \citep{cubuk2018autoaugment} (Section \ref{autoaug}) , and can improve COCO object detection performance as well (Section \ref{object-detection}).
    \item We perform a frequency-based analysis~\cite{yin2019fourier} of models trained with \aug{Patch Gaussian} and find that they can better leverage high-frequency information in lower layers, while not being too sensitive to them at later ones (Section \ref{section-fourier-analysis}).
\end{itemize}

\section{Preliminaries}

We start by considering two data augmentations: \aug{Cutout} \cite{devries2017improved} and \aug{Gaussian} \cite{grandvalet1997noise}.
The former sets a random patch of the input images to a constant (the mean pixel in the dataset)
and is successful at improving clean accuracy. The latter works by adding independent Gaussian noise to each pixel of the input image, which can increase robustness to Gaussian noise directly.

To apply \aug{Gaussian}, we uniformly sample a standard deviation $\sigma$ from 0 up to some maximum value $\sigma_{max}$, and add i.i.d. noise sampled from $\mathcal{N}(0, \sigma^2)$ to each pixel. To apply \aug{Cutout}, we use a fixed patch size $W$, and randomly set a square region with size $W\times W$ to the constant mean of each RGB channel in the dataset.
As in~\cite{devries2017improved}, the patch location is randomly sampled and can lie outside of the $32\times 32$ CIFAR-10 (or $224\times 224$ ImageNet) image but its center is constrained to lie within it.
Patch sizes and $\sigma_{max}$ are selected based on the method described in Section \ref{hyper-selection}.

\subsection{Cutout and Gaussian exhibit a trade-off between accuracy and robustness}
\label{section-tradeoff}

\begin{figure}[t]
\begin{center}
\begin{minipage}[r]{0.4\textwidth}
\hfill\includegraphics[trim={0 0 15em 0},clip=true,height=11.5em]{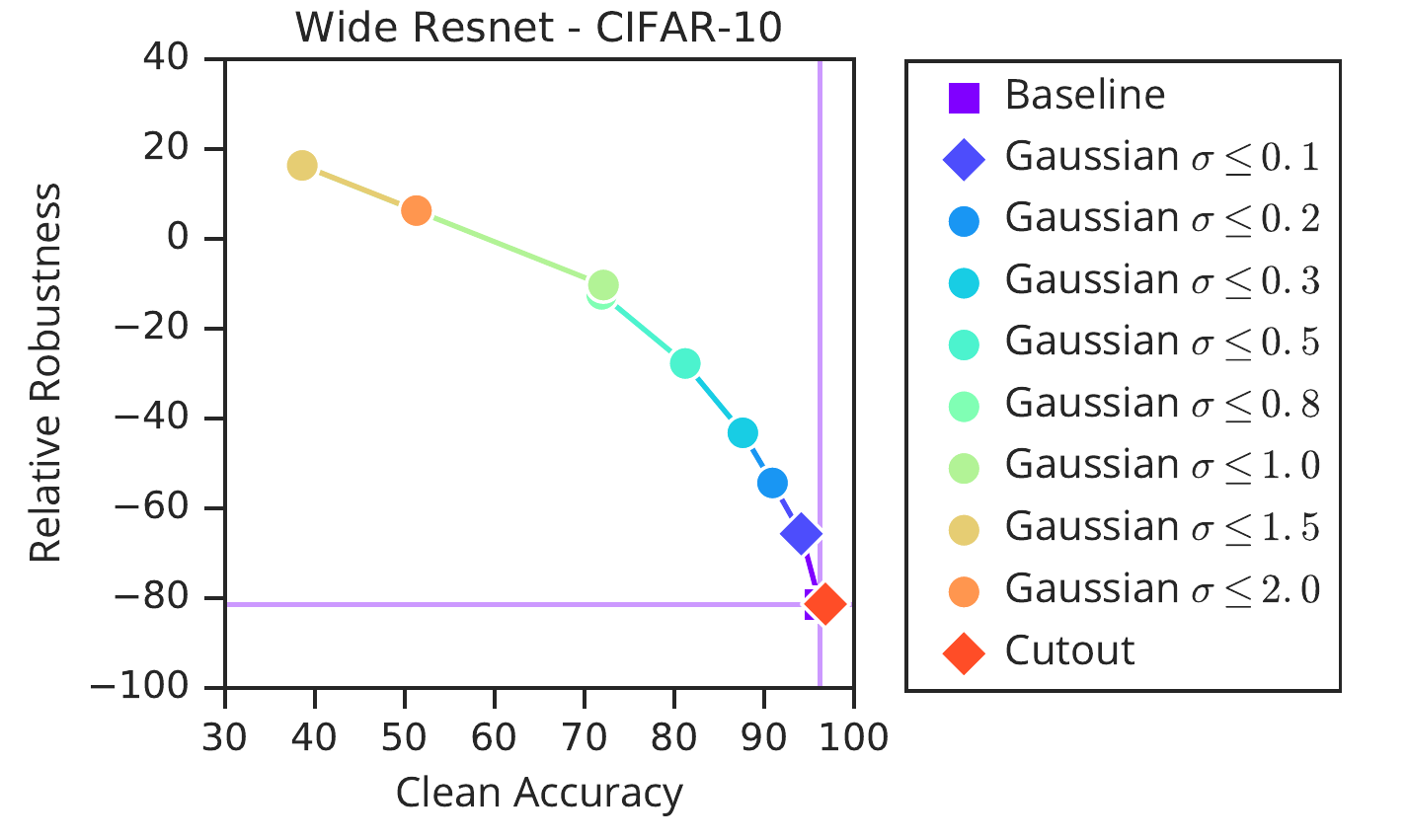}
\end{minipage}%
\begin{minipage}[l]{0.6\textwidth}
\includegraphics[height=11.5em]{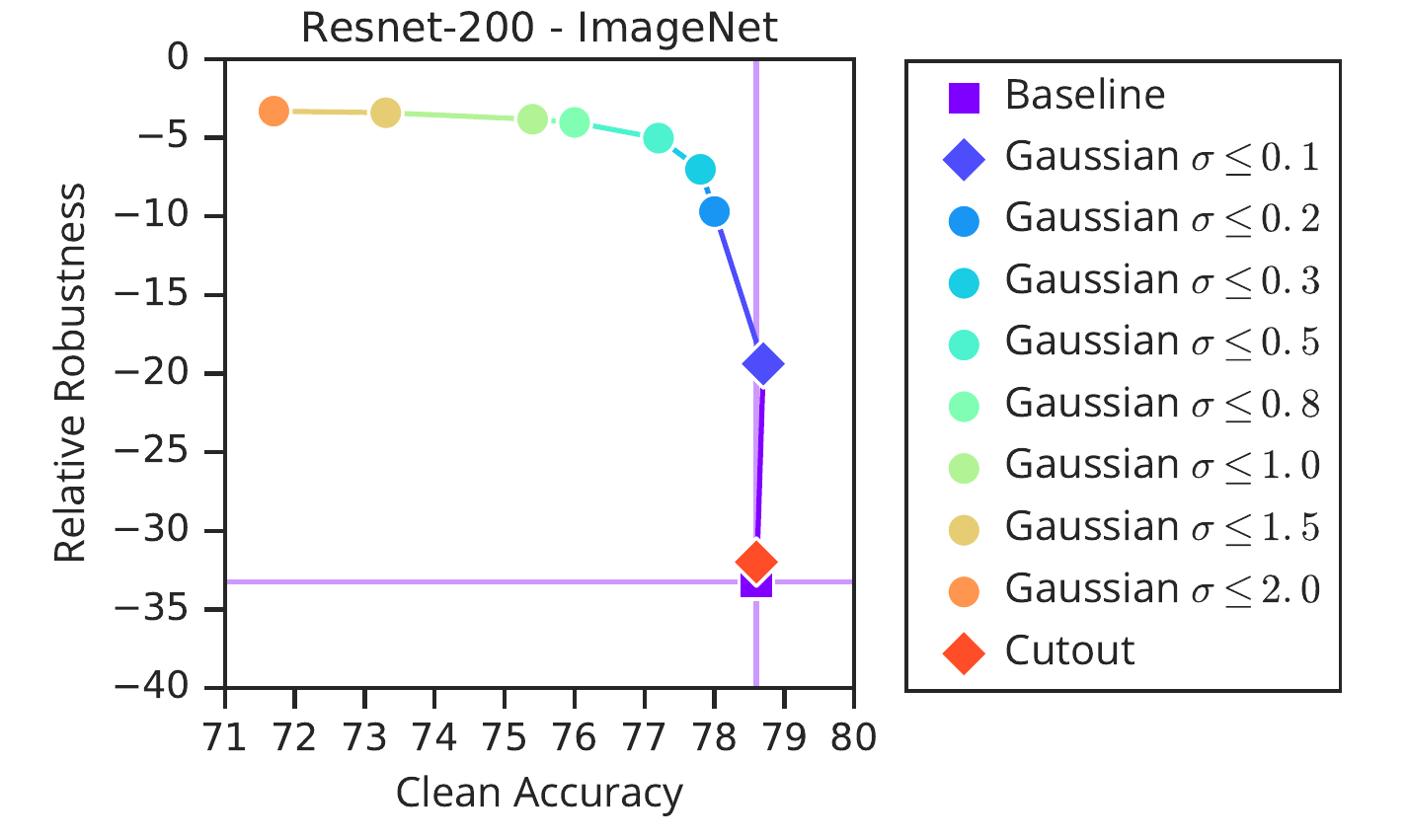}\hfill
\end{minipage}
\caption{
The robustness-accuracy trade-off between \aug{Cutout} and \aug{Gaussian} augmentations.
Each dot represents a model trained with different augmentations and hyper-parameters. The y-axis is the change in accuracy when tested on data corrupted with Gaussian noise at various $\sigma$ (average corrupted accuracy minus clean accuracy). The diamond indicates augmentation hyper-parameters selected by the method in Section \ref{hyper-selection}.
}
\label{cutout-vs-fullgauss}
\end{center}
\end{figure}

We compare the effectiveness of \aug{Gaussian} and \aug{Cutout} data augmentation for accuracy and robustness by measuring the performance of models trained with each on clean data, as well as data corrupted by various standard deviations of Gaussian noise. 
Figure~\ref{cutout-vs-fullgauss} highlights an apparent trade-off in using these methods. In accordance to previous work \cite{devries2017improved}, \aug{Cutout} improves accuracy on clean test data. Despite this, we find it does not lead to increased robustness. Conversely, training with higher $\sigma$
of \aug{Gaussian} can lead to increased robustness to Gaussian noise, but it also leads to decreased accuracy on clean data. Therefore, any robustness gains are offset by poor overall performance.

At first glance, these results seem to reinforce the findings of previous work \citep{tsipras2018robustness}, indicating that robustness comes at the cost of generalization. In the following sections, we will explore whether there exists augmentation strategies that do not exhibit this limitation.
\section{Method}

Each of the two methods seen so far achieves one half of our stated goal: either improving robustness or improving clean test accuracy, but never both. 
To overcome the limitations of existing data augmentation techniques, we introduce \aug{Patch Gaussian}, a new technique that combines the noise robustness of \aug{Gaussian} with the improved clean accuracy of \aug{Cutout}.

\subsection{Patch Gaussian}
\label{section-patch-gauss}

\begin{wrapfigure}{r}{0.48\textwidth}%
\vspace{-0.35in}
\begin{center}%
\begin{minipage}[c]{0.48\textwidth}%
\centerline{\includegraphics[width=\textwidth]{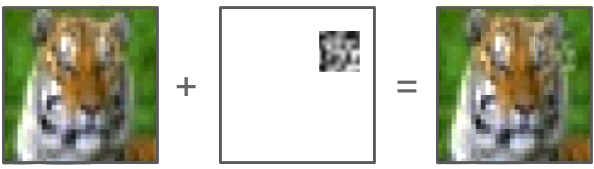}}%
\end{minipage}%
\caption{\aug{Patch Gaussian}
is the addition of Gaussian noise to pixels in a square patch.
It allows us to interpolate between \aug{Gaussian} and \aug{Cutout}, approaching \aug{Gaussian} with increasing patch size and \aug{Cutout} with increasing $\sigma$.
}%
\label{patch-gauss-method}%
\end{center}
\vspace{-2em}
\end{wrapfigure} 

\begin{figure}[ht]
\begin{center}
\centerline{
    \begin{tabular}{lcccccc}
    & \multicolumn{6}{c}{Patch Size $W$}\\
    & $=20$ & $=30$ & $=50$ & $=100$ & $=150$ & $=448$\\
    \raisebox{2.5\normalbaselineskip}[0pt][0pt]{\rotatebox[origin=c]{90}{$\sigma=0.1$}} & \includegraphics[width=0.15\textwidth]{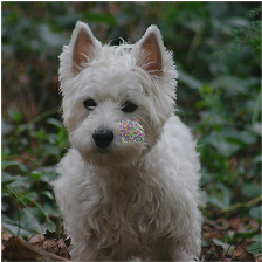} & \includegraphics[width=0.15\textwidth]{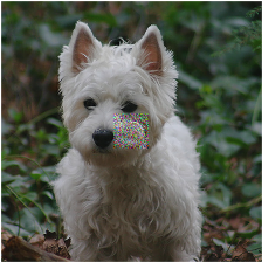} & \includegraphics[width=0.15\textwidth]{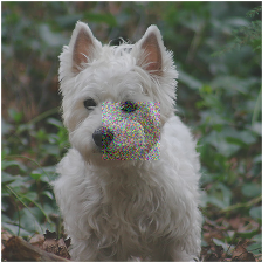} & \includegraphics[width=0.15\textwidth]{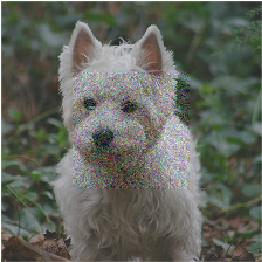} & \includegraphics[width=0.15\textwidth]{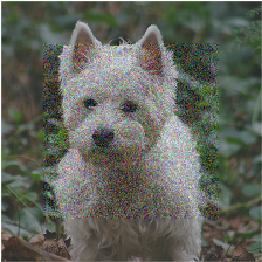} & \includegraphics[width=0.15\textwidth]{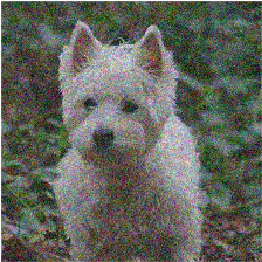}\\
    \raisebox{2.5\normalbaselineskip}[0pt][0pt]{\rotatebox[origin=c]{90}{$\sigma=0.2$}} & \includegraphics[width=0.15\textwidth]{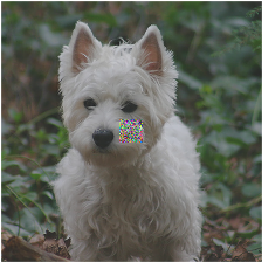} & \includegraphics[width=0.15\textwidth]{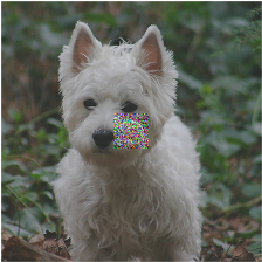} & \includegraphics[width=0.15\textwidth]{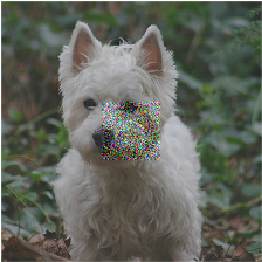} & \includegraphics[width=0.15\textwidth]{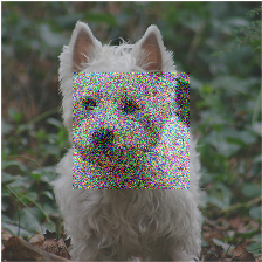} & \includegraphics[width=0.15\textwidth]{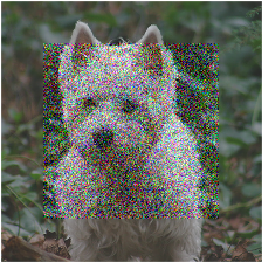} & \includegraphics[width=0.15\textwidth]{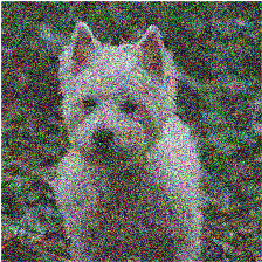}\\
    \raisebox{2.5\normalbaselineskip}[0pt][0pt]{\rotatebox[origin=c]{90}{$\sigma=0.3$}} & \includegraphics[width=0.15\textwidth]{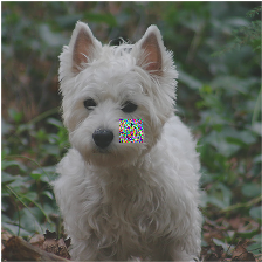} & \includegraphics[width=0.15\textwidth]{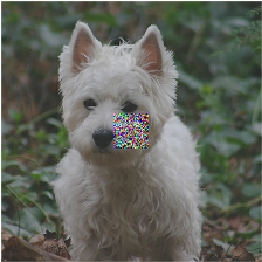} & \includegraphics[width=0.15\textwidth]{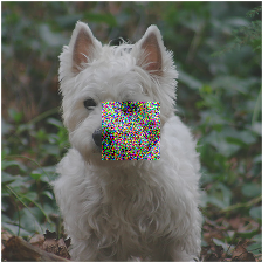} & \includegraphics[width=0.15\textwidth]{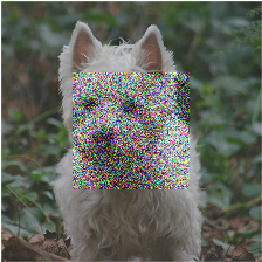} & \includegraphics[width=0.15\textwidth]{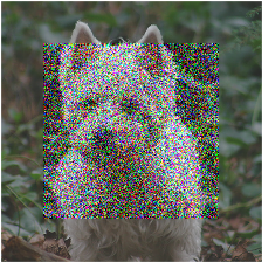} & \includegraphics[width=0.15\textwidth]{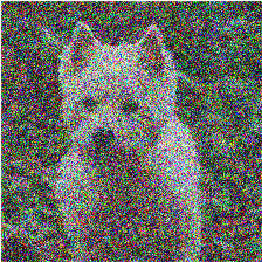}\\
    \raisebox{2.5\normalbaselineskip}[0pt][0pt]{\rotatebox[origin=c]{90}{$\sigma=0.5$}} & \includegraphics[width=0.15\textwidth]{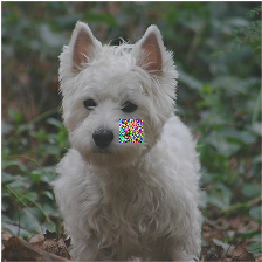} & \includegraphics[width=0.15\textwidth]{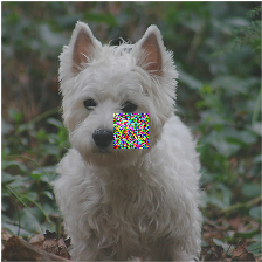} & \includegraphics[width=0.15\textwidth]{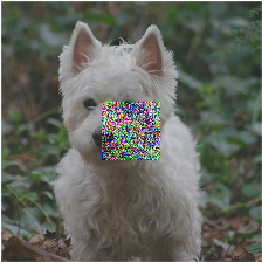} & \includegraphics[width=0.15\textwidth]{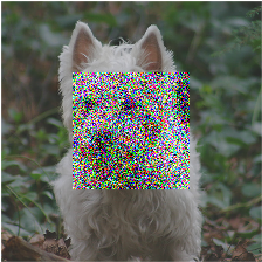} & \includegraphics[width=0.15\textwidth]{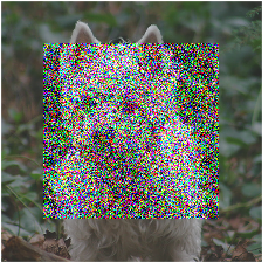} & \includegraphics[width=0.15\textwidth]{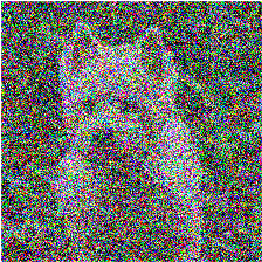}\\
    \raisebox{2.5\normalbaselineskip}[0pt][0pt]{\rotatebox[origin=c]{90}{$\sigma=0.8$}} & \includegraphics[width=0.15\textwidth]{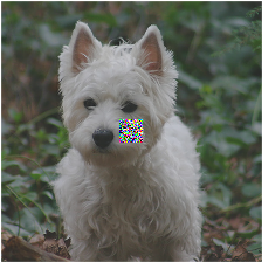} & \includegraphics[width=0.15\textwidth]{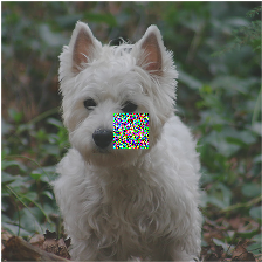} & \includegraphics[width=0.15\textwidth]{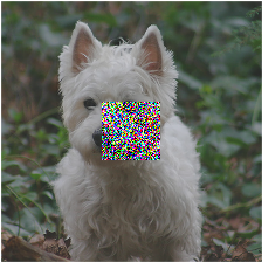} & \includegraphics[width=0.15\textwidth]{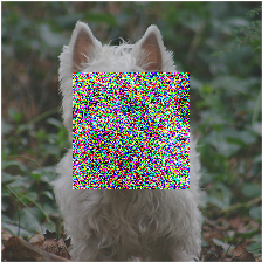} & \includegraphics[width=0.15\textwidth]{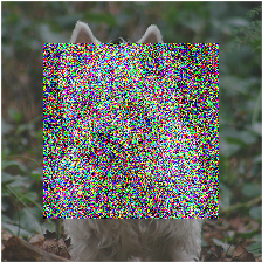} & \includegraphics[width=0.15\textwidth]{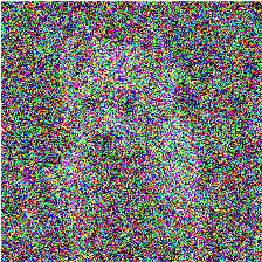}\\
    \raisebox{2.5\normalbaselineskip}[0pt][0pt]{\rotatebox[origin=c]{90}{$\sigma=1.0$}} & \includegraphics[width=0.15\textwidth]{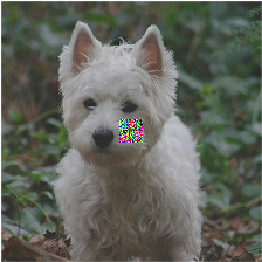} & \includegraphics[width=0.15\textwidth]{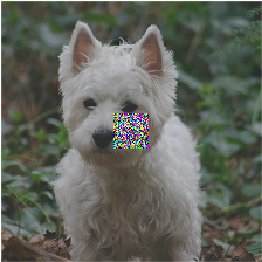} & \includegraphics[width=0.15\textwidth]{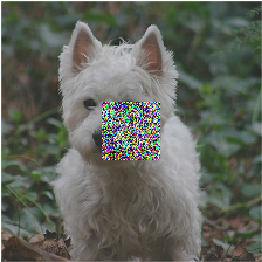} & \includegraphics[width=0.15\textwidth]{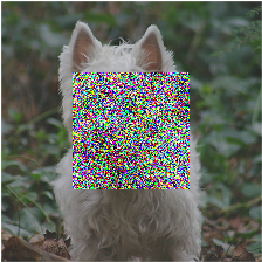} & \includegraphics[width=0.15\textwidth]{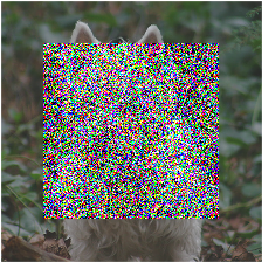} & \includegraphics[width=0.15\textwidth]{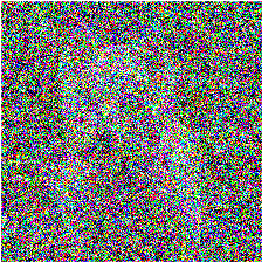}\\
    \end{tabular}
}
\caption{Images modified with \aug{Patch Gaussian}, with centered patch, at various $W$\,\&\,$\sigma$.
}
\label{patch-gauss-image-viz}
\end{center}
\end{figure}

\aug{Patch Gaussian} works by adding
a $W$x$W$ patch of Gaussian noise to the image (Figure \ref{patch-gauss-method})\footnote{A TensorFlow implementation of \aug{Patch Gaussian} can be found in Appendix (Figure \ref{implementation}).}. As with \aug{Cutout}, the center of the patch is sampled to be within the image. %
By varying the size of this patch and the maximum standard deviation of noise sampled $\sigma_{max}$, we can interpolate between \aug{Gaussian} (which applies additive Gaussian noise to the whole image) and an approximation of Cutout (which removes all information inside the patch). See Figure~\ref{patch-gauss-image-viz} for more examples.

All image transformations, including \aug{Patch Gaussian} are performed on images with unnormalized pixel values in $[0,1]$ range.
For all images, standard random flipping and cropping is applied immediately \textit{after} any augmentations mentioned on CIFAR-10 (\textit{before}, on Imagenet). After noise-based augmentations, images are clipped to the $[0,1]$ range.

\subsection{Hyper-parameter selection}
\label{hyper-selection}
Our goal is to learn models that achieve both good clean accuracy and improved robustness to corruptions.
When selecting hyper-parameters we need to decide how to weight these two metrics.
Here, we focused on identifying the models that were most robust while still achieving a minimum accuracy (Z) on the clean test data. 
Values of Z vary per dataset and model, and can be found in the Appendix (Table~\ref{hyper-table}). If no model has clean accuracy $\geq$Z, we report the model with highest clean accuracy, unless otherwise specified. We find that patch sizes around $25$ on CIFAR ($\leq$250 on ImageNet, i.e.: uniformly sampled with maximum value 250) with $\sigma\leq1.0$ generally perform the best. A complete list of selected hyper-parameters for all augmentations can be found in Table~\ref{hyper-table}.

Here, robustness is defined as average accuracy of the model, when tested on data corrupted by various $\sigma$ ($0.1$, $0.2$, $0.3$, $0.5$, $0.8$, $1.0$) of Gaussian noise, relative to the clean accuracy. 
This metric is correlated with mCE \cite{ford2019adversarial}, so it ensures model rosbustness is generally useful beyond Gaussian corruptions.
By picking models based on their Gaussian noise robustness, we ensure that our selection process does not overfit to the Common Corruptions benchmark \cite{hendrycks2018benchmarking}.%
\[\text{Relative Gaussian Robustness} = \expectation_\sigma(\text{Accuracy on Data Corrupted by }\sigma) - \text{Clean Accuracy}\]

\subsection{Models, Datasets, \& Implementation Details}
\label{models-datasets}

We run our experiments on CIFAR-10 \cite{krizhevsky2009learning} and ImageNet \cite{imagenet2009} datasets. On CIFAR-10, we use the Wide-ResNet-28-10 model \cite{zagoruyko2016wide}, as well as the Shake-shake-112 model \cite{gastaldi2017shake}, trained for 200 epochs and 600 epochs respectively. The Wide-ResNet model uses a initial learning rate of 0.1 with a cosine decay schedule. Weight decay is set to be $5$e-$4$ and batch size is 128. We train all models, including the baseline, with standard data augmentation of horizontal flips and pad-and-crop. 
Our code uses the same hyper parameters as \cite{cubuk2018autoaugment} \footnote{Available at \texttt{https://github.com/tensorflow/models/tree/master/research/autoaugment}}.

On ImageNet, we use the ResNet-50 and Resnet-200 models \cite{he2016deep}, trained for 90 epochs. We use a weight decay rate of $1$e-$4$, global batch size of 512 and learning rate of 0.2. The learning rate is decayed by 10 at epochs 30, 60, and 80. We use standard data augmentation of horizontal flips and crops. All CIFAR-10 and ImageNet experiments use the listed hyper-parameters above, unless specified otherwise.
Our code uses the same hyper parameters as open-sourced implementations\footnote{Available at \texttt{https://github.com/tensorflow/tpu/tree/master/models/official/resnet}}.

\section{Results}
We show that models trained with \aug{Patch Gaussian} can overcome the trade-off observed in Fig.~\ref{cutout-vs-fullgauss} and learn models that are robust while maintaining their generalization accuracy (Section \ref{section-overcoming}). In doing so, we establish a new state of the art in CIFAR-C and ImageNet-C Common Corruptions benchmark \cite{hendrycks2018benchmarking} (Section \ref{section-sota-on-cc}). We then show that \aug{Patch Gaussian} can be used in complement to other common regularization strategies (Section \ref{compare-other-regs}), data augmentation policies~\cite{cubuk2018autoaugment} (Section \ref{autoaug}), and that it can also improve training of object detection models (Section \ref{object-detection})

\subsection{Patch Gaussian overcomes this trade-off and improves both accuracy and robustness}
\label{section-overcoming}
\begin{figure}[b]
\begin{center}
\begin{minipage}[c]{0.5\textwidth}
\centerline{\includegraphics[width=\textwidth]{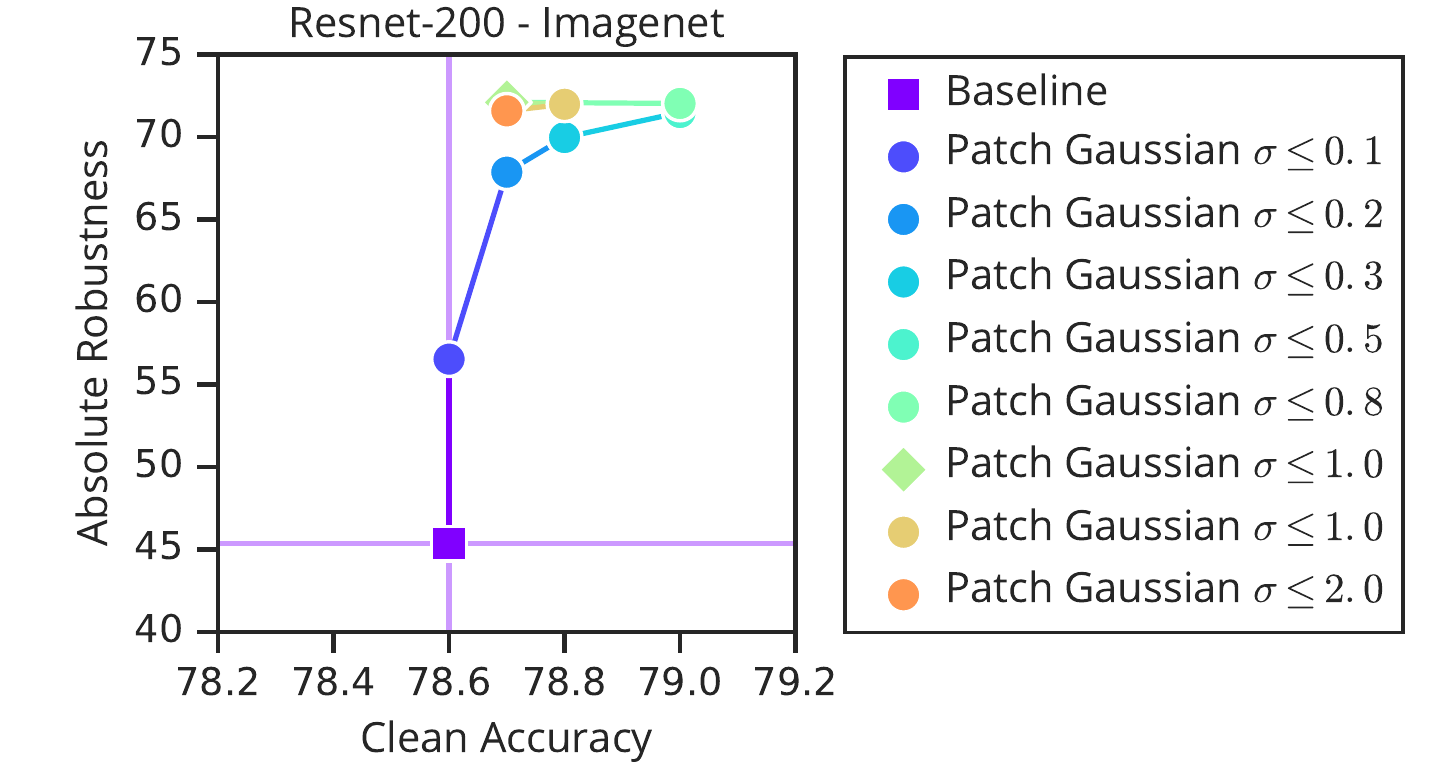}}
\end{minipage}%
\begin{minipage}[c]{0.5\textwidth}
\centerline{\includegraphics[width=\textwidth]{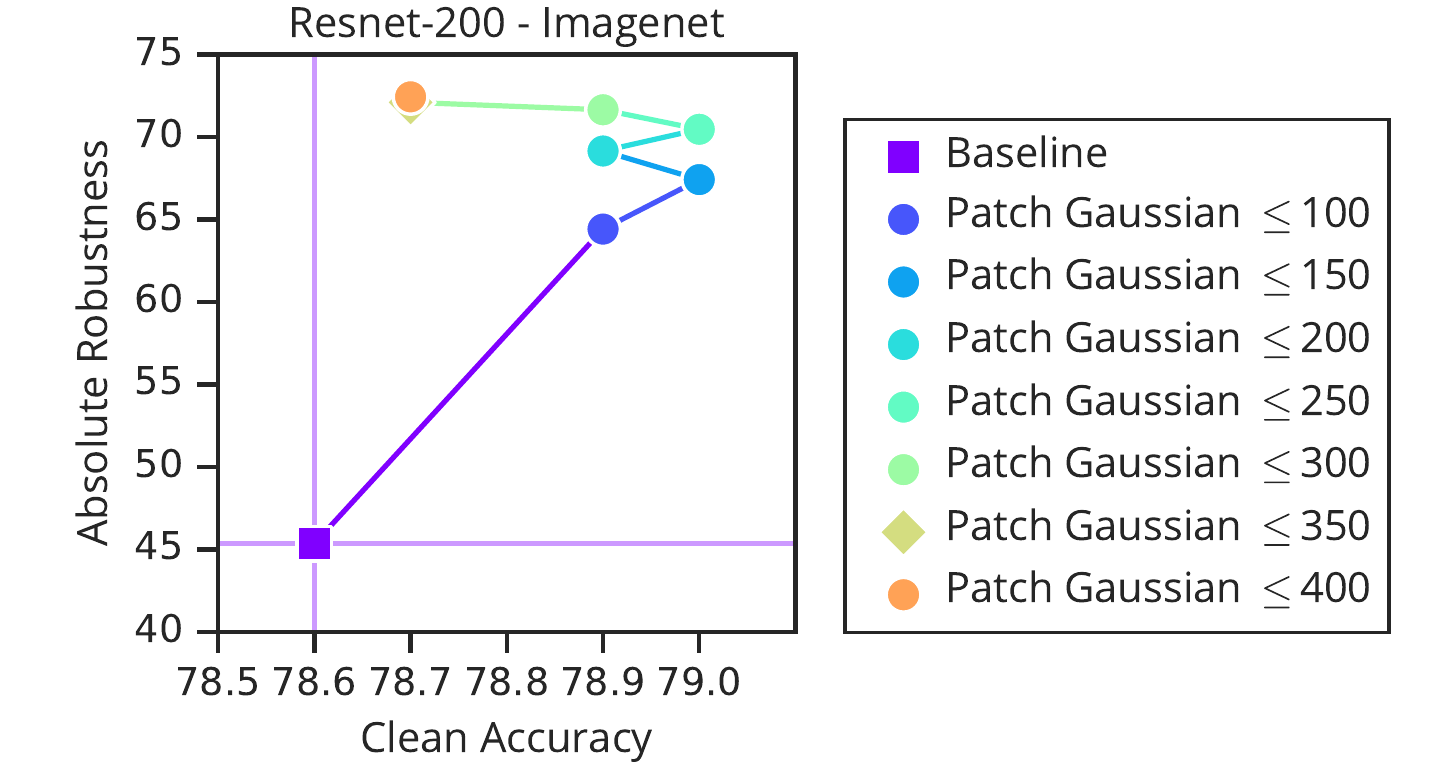}}
\end{minipage}%
\caption{Training with \aug{Patch Gaussian} improves clean data accuracy and robustness simultaneously. Each dot represents a model trained with various $\sigma$ (left) or patch sizes (right), while keeping the other fixed at the value indicated by the diamond. The y-axis is the average absolute accuracy when tested on data corrupted by Gaussian noise at various $\sigma$. The diamond indicates the augmentation hyper-parameters selected by the method in Section \ref{hyper-selection}.
}
\label{patch-gauss-overcoming}
\end{center}
\end{figure}

We train models on various hyper-parameters of \aug{Patch Gaussian} and find that the model selected by the method in Section \ref{hyper-selection} leads to improved robustness to Gaussian noise, like \aug{Gaussian}, while also improving clean accuracy, much like \aug{Cutout}. In Figure \ref{patch-gauss-overcoming}, we visualize these results in an ablation study, either varying patch sizes and fixing $\sigma$ to the selected value ($1.0$) or varying noise level $\sigma$ and fixing the patch size to the selected value ($350$).

\subsection{Training with Patch Gaussian leads to improved Common Corruption robustness}
\label{section-sota-on-cc}

In this section, we look at how our augmentations impact robustness in a more realistic setting beyond robustness to additive Gaussian noise. Rather than focusing on adversarial examples that are worst-case bounded perturbations, we focus on a more general set of corruptions~\cite{gilmer2018motivating} that models are likely to encounter in real-world settings: the Common Corruptions benchmark \cite{hendrycks2018benchmarking}.
This benchmark, also referred to as CIFAR-C and ImageNet-C, is composed of images transformed with 15 corruptions, at 5 severities each. The corruptions are designed to model those commonly found in real-world settings, such as brightness, different weather conditions, and different kinds of noise.

Tables \ref{cifar-table} and \ref{imagenet-table} show that \aug{Patch Gaussian} achieves state of the art on both of these benchmarks in terms of 
mean Corruption Error (mCE). However, ImageNet-C was released in compressed JPEG format \cite{jpeg}, which alters the corruptions applied to the raw pixels. Therefore, we report results on the benchmark as-released (``Original mCE'') as well as a version of 12 corruptions without the extra compression (``mCE'')
\footnote{Available at \texttt{https://github.com/tensorflow/datasets} under \texttt{imagenet2012\_corrupted}}.

\begin{table}[h]
  \caption{\aug{Patch Gaussian} achieves state of the art in the CIFAR-C benchmark \cite{hendrycks2018benchmarking} while improving clean accuracy.
Augmentation hyper-parameters were selected based on the method in Section \ref{hyper-selection} and can be found in Appendix.
*\aug{Cutout 16} is presented for direct comparison with  \cite{devries2017improved,gastaldi2017shake}.} 
  \label{cifar-table}
  \centering
  \begin{tabular}{llccc}
    \toprule
        & Augmentation & Test Accuracy & mCE & mCE (-noise) \\
    \midrule
    \multirow{6}{*}{\rotatebox[origin=c]{90}{\linebreakcell{Wide\\Resnet-28-10}}} & Adversarial & 87.3\% & 1.049 & 1.157\\
        \cmidrule(r){2-5}
        & Baseline & 96.2\% & 1.000 & 1.000\\
        & Cutout & 96.8\% & 1.265 & 1.185\\
        & Cutout 16* & \textbf{97.0\%} & 1.002 & 0.953\\
        & Gaussian & 94.1\% & 0.887 & 0.995\\
        & Patch Gaussian & 96.6\% & \textbf{0.797} & \textbf{0.858}\\
    \midrule
    \multirow{5}{*}{\rotatebox[origin=c]{90}{Shake 112}} & Baseline & 96.8\% & 1.000 & 1.000\\
        & Cutout & 97.1\% & 0.946 & 0.930\\
        & Cutout 16* & \textbf{97.5\%} & 0.912 & 0.872\\
        & Gaussian & 94.6\% & 0.977 & 1.111\\
        & Patch Gaussian & 97.2\% & \textbf{0.713} & \textbf{0.776}\\
    \bottomrule
  \end{tabular}
\end{table}

\begin{table}[h]
  \caption{\aug{Patch Gaussian} achieves state of the art in the ImageNet-C benchmark \cite{hendrycks2018benchmarking} while improving uncorrupted test accuracy. ``SIN+IN fIN'' is the shape-biased model from~\cite{geirhos2018imagenet}. ``Original mCE'' refers to the jpeg-compressed benchmark, as used in \cite{geirhos2018imagenet, hendrycks2018benchmarking}. ``mCE'' is a version of it without the extra jpeg compression. Note that \aug{Patch Gaussian} improves robustness even in corruptions that aren't noise-based. Augmentation hyper-parameters were selected based on the method in Section \ref{hyper-selection} and can be found in Appendix. For Resnet-200, we also present \aug{Gaussian} at a higher $\sigma$ to highlight the accuracy-robustness trade-off.
  }
  \label{imagenet-table}
  \centering
  \centerline{
  \begin{tabular}{llccccc}
    \toprule
        & Augmentation & Test Accuracy & Original mCE & Original mCE (-noise) & mCE & mCE (-noise) \\
    \midrule
    \multirow{5}{*}{\rotatebox[origin=c]{90}{Resnet-50}} & SIN+IN ftIN & 76.7\% & 0.738 & 0.731 & - & - \\
        \cmidrule(r){2-7}
        & Baseline  & \textbf{76.4\%} & 0.753 & 0.763 & 1.00 & 1.00\\
        & Cutout  & 76.2\% & 0.758 & 0.766 & 1.007 & 1.005\\
        & Gaussian & 75.6\% & 0.739 & 0.754 & 0.898 & 0.991\\
        & Patch Gaussian & 76.0\% & \textbf{0.714} & \textbf{0.736} & \textbf{0.872} & \textbf{0.969}\\
    \midrule
    \multirow{5}{*}{\rotatebox[origin=c]{90}{Resnet-200}} & Baseline & 78.6\% & 0.675 & 0.686 & 0.881 & 0.883 \\
        & Cutout & 78.6\% & 0.671 & 0.687 & 0.874 & 0.884\\
        & Gaussian & \textbf{78.7\%} & 0.658 & 0.678 & 0.795 & 0.881\\
        & Gaussian ($\sigma\leq0.2$) & 78.1\% & 0.644 & 0.665 & 0.784 & 0.874\\
        & Patch Gaussian & \textbf{78.7\%} & \textbf{0.604} & \textbf{0.634} & \textbf{0.736} & \textbf{0.818}\\
    \bottomrule
  \end{tabular}
  }
\end{table}

To compute mCE, we normalize the corruption error for each model and dataset to the baseline with only flip and crop data augmentation.
The one exception is Original mCE ImageNet, where we use the AlexNet baseline to be directly comparable with previous work \cite{hendrycks2018benchmarking,geirhos2018imagenet}.

Because \aug{Patch Gaussian} is a noise-based augmentation, we wanted to verify whether its gains on this benchmark were solely due to improved performance on noise-based corruptions (Gaussian Noise, Shot Noise, and Impulse Noise). To do this, we also measure the models' average performance on all \textit{other} corruptions, reported as ``Original mCE (-noise)'', and ``mCE (-noise)''. We observe that \aug{Patch Gaussian} outperforms all other models, even on corruptions like fog where \aug{Gaussian} hurts performance \cite{ford2019adversarial}. 
Scores for each corruption can be found in the Appendix (Tables \ref{full-og-cc} and \ref{full-cc}).

Comparing the lower capacity ResNet-50 and Wide ResNet models to higher-capacity ResNet-200 and Shake 112 models, we find diminished gains in clean accuracy and robustness. 
Still, \aug{Patch Gaussian} achieves a substantial increase in mCE relative to other augmentation strategies.

\subsection{Patch Gaussian can be combined with other regularization strategies}
\label{compare-other-regs}
Since \aug{Patch Gaussian} has a regularization effect on the models trained above, we compare it with other regularization methods: larger weight decay, label smoothing, and dropblock (Table~\ref{tab:regularization}). We find that while label smoothing improves clean accuracy, it weakens the robustness in all corruption metrics we have considered. This agrees with the theoretical prediction from \cite{cubuk2017intriguing}, which argued that increasing the confidence of models would improve robustness, whereas label smoothing reduces the confidence of predictions. We find that increasing the weight decay from the default value used in all models does not improve clean accuracy or robustness.

Here, we focus on analyzing the interaction of different regularization methods with \aug{Patch Gaussian}.
Previous work indicates that improvements on the clean accuracy appear after training with Dropblock for 270 epochs~\cite{ghiasi2018dropblock}, but we did not find that training for 270 epochs changed our analysis. Thus, we present models trained at 90 epochs for direct comparison with other results.
Due to the shorter training time, Dropblock does not improve clean accuracy, yet it does make the model more robust (relative to baseline) according to all corruption metrics we consider. %

We find that using label smoothing in addition to \aug{Patch Gaussian} has a mixed effect, it improves clean accuracy while slightly improving robustness metrics except for the Original mCE. Combining Dropblock with \aug{Patch Gaussian} reduces the clean accuracy relative to the \aug{Patch Gaussian}-only model, as Dropblock seems to be a strong regularizer when used for 90 epochs. However, using Dropblock and \aug{Patch Gaussian} together leads to the best robustness performance. These results indicate that \aug{Patch Gaussian} can be used in conjunction with existing regularization strategies.

\begin{table}[h]
  \vspace{0.3in}
  \caption{\aug{Patch Gaussian} can be used with other regularization methods for improved robustness. ``Original mCE'' refers to the jpeg-compressed benchmark, as used in \cite{geirhos2018imagenet, hendrycks2018benchmarking}. ``mCE'' is a version of it without the extra jpeg compression. All of the models are ResNet-50 trained on ImageNet with same hyperparameters for 90 epochs.
}
  \label{tab:regularization}
  \centering
  \centerline{
  \begin{tabular}{llccccc}
    \toprule
        & Regularization & \linebreakcell{Test\\Accuracy} & \linebreakcell{Original\\mCE} & \linebreakcell{Original\\mCE (-noise)} & mCE & mCE (-noise) \\
    \midrule
        \multirow{5}{*}{\rotatebox[origin=c]{90}{Resnet-50}} & Label Smoothing &           \textbf{76.7\%} & 0.760 & 0.765 & 1.01  & 1.01\\
        & Larger Weight Decay ($0.001$) & 74.9\%  & 0.766 & 0.777 & 1.02  & 1.03 \\
        & Dropblock &                         76.3\%  & 0.734 & 0.743 & 0.971 & 0.974 \\
        & Patch Gaussian + Label Smoothing & 76.5\% & 0.720 & 0.734 &\textbf{0.868} & 0.966\\
        & Patch Gaussian + Dropblock & 75.7\% & \textbf{0.708} & \textbf{0.726} & 0.870 & \textbf{0.961}\\
    \bottomrule
  \end{tabular}
  }
\end{table}

\newpage
\subsection{Patch Gaussian can be combined with AutoAugment policies for improved results}
\label{autoaug}

Knowing that \aug{Patch Gaussian} can be combined with other regularizers, it's natural to ask whether it can also be combined with other data augmentation policies. Table \ref{tab:autoaugment} highlights models trained with AutoAugment \cite{cubuk2018autoaugment}. For fair comparison of mCE scores, we train all models with the best AutoAugment policy, but without contrast and Inception color pre-processing, as those are present in the Common Corruptions benchmark. 
This process is imperfect since AutoAugment has \emph{many} operations, some of which could still be correlated with corruptions.
Regardless, we find that \aug{Patch Gaussian} improves accuracy and robustness over simply using AutoAugment. 

Because AutoAugment leads to state of the art accuracies, we are interested in seeing how far it can be combined with \aug{Patch Gaussian} to improve results. Therefore, and unlike previous experiments, models are trained for 180 epochs to yield best results possible.

\begin{table}[h]
  \caption{\aug{Patch Gaussian} can be combined with AutoAugment~\cite{cubuk2018autoaugment} data augmentation policy for improved results. ``Original mCE'' refers to the jpeg-compressed benchmark, as used in \cite{geirhos2018imagenet, hendrycks2018benchmarking}. ``mCE'' is a version of it without the extra jpeg compression. All of the models are ResNet-50 trained on ImageNet with best AutoAugment policy for 180 epochs, to highlight improvements.
}
  \label{tab:autoaugment}
  \centering
  \centerline{
  \begin{tabular}{llccccc}
    \toprule
        & \leftlinebreakcell{Model\\(trained with AutoAugment)} & \linebreakcell{Test\\Accuracy} & \linebreakcell{Original\\mCE} & \linebreakcell{Original\\mCE (-noise)} & mCE & mCE (-noise) \\
    \midrule
        \multirow{2}{*}{\rotatebox[origin=c]{90}{\linebreakcell{Res\\Net\\50}}} & Baseline & 77.0\% & 0.674 & 0.697 & 0.855  & 0.882\\
        & Patch Gaussian ($W$=$150, \sigma\le0.5$) & \textbf{77.3\%}  & \textbf{0.656} & \textbf{0.682} & \textbf{0.779}  & \textbf{0.863} \\
    \bottomrule
  \end{tabular}
  }
\end{table}

\subsection{Patch Gaussian improves performance in object detection}
\label{object-detection}
Since \aug{Patch Gaussian} can be combined with both regularization strategies as well as data augmentation policies, we want to see if it is generally useful beyond classification tasks.
We train a RetinaNet detector~\cite{lin2017focal} with ResNet-50 backbone~\cite{he2016deep} on the COCO dataset~\cite{lin2014microsoft}. Images for both baseline and \aug{Patch Gaussian} models are horizontally flipped half of the time, after being resized to $640 \times 640$. We train both models for 150 epochs using a learning rate of 0.08 and a weight decay of $1e-4$. The focal loss parameters are set to be $\alpha=0.25$ and $\gamma=1.5$.

Despite being designed for classification, \aug{Patch Gaussian} improves detection performance according to all metrics when tested on the clean COCO validation set (Table~\ref{tab:detection}). On the primary COCO metric mean average precision (mAP), the model trained with \aug{Patch Gaussian} achieves a 1\% higher accuracy over the baseline, whereas the model trained with \aug{Gaussian} suffers a 2.9\% loss.%

\begin{table}[b]
  \caption{Mean average precision (mAP) on COCO with baseline augmentation of horizontal flips and \aug{Patch Gaussian}. mAP$_{\texttt S}$, mAP$_{\texttt M}$, and mAP$_{\texttt L}$ refer to mAP for small, medium, and large objects, respectively. mAP$_{\texttt 50}$ and mAP$_{\texttt 75}$ refer to mAP at intersection over union values of 50 and 75, respectively. mAP in the final column is the averaged mAP over IoU=0.5:0.05:0.95.      }
  \label{tab:detection}
  \centerline{
\begin{tabular}{c|l|rrrrr|r}
  \toprule
  Tested on & & mAP$_{\texttt S}$ &  mAP$_{\texttt M}$ &  mAP$_{\texttt L}$ & mAP$_{\texttt 50}$ &  mAP$_{\texttt 75}$ &   mAP  \\
  \midrule
  \multirow{3}{*}{\linebreakcell{Clean\\Data}} & Baseline  & 15.6&36.9& 48.3 & 50.8 &35.6& 33.2  \\
  & Gaussian ($\sigma\le1.0$) & 13.1& 32.6& 44.0& 45.7 &31.2  & 29.3  \\
  & Patch Gaussian ($W$=$200, \sigma\le1.0$) & \textbf{16.1}&\textbf{37.9}& \textbf{50.3} &\textbf{ 51.9} &\textbf{36.5}&\textbf{34.2}  \\
  \midrule
  \multirow{3}{*}{\linebreakcell{Gaussian Noise\\(=$0.25$)}}%
  & Baseline  & 4.5 & 12.7 & 17.6 & 19.3 & 11.7 & 11.6\\
  & Gaussian ($\sigma\le1.0$) & 9.9& 28.1& \textbf{41.0} & \textbf{41.7} & 26.8 & \textbf{26.1}  \\
  & Patch Gaussian ($W$=$200, \sigma\le1.0$) & \textbf{10.1} & \textbf{28.2} & 40.4 & 41.3 & \textbf{27.2} & \textbf{26.1}  \\
  \bottomrule
\end{tabular}
  }
\end{table}

Next, we evaluate these models on the validation set corrupted by i.i.d. Gaussian noise, with $\sigma=0.25$.
We find that model trained with \aug{Gaussian} and \aug{Patch Gaussian} achieve the highest mAP of 26.1\% on the corrupted data, whereas the baseline achieves 11.6\%. It is interesting to note that \aug{Patch Gaussian} model achieves a better result on the harder metrics of small object detection and stricter intersection over union (IOU) thresholds, whereas the \aug{Gaussian} model achieves a better result on the easier tasks of large object detection and less strict IOU threshold metric.

Overall, as was observed for the classification tasks, training object detection models with \aug{Patch Gaussian} leads to significantly more robust models without sacrificing clean accuracy.

\section{Discussion}

In an attempt to understand \aug{Patch Gaussian}'s performance, we perform a frequency-based analysis of models trained with various augmentations using the method introduced in~\cite{yin2019fourier}.

First, we perturb each image in the dataset with noise sampled at each orientation and frequency in Fourier space. Then, we measure changes in the network activations and test error when evaluated with these Fourier-noise-corrupted images: we measure the change in $\ell_2$ norm of the tensor directly after the first convolution, as well as the absolute test error. This procedure yields a heatmap, which indicates model sensitivity to different frequency and orientation perturbations in the Fourier domain. 
Each image in Fig \ref{fourier} shows first layer (or test error) sensitivity as a function of frequency and orientation of the sampled noise, with the middle of the image containing the lowest frequencies, and the edges of the image containing highest frequencies.

For CIFAR-10 models, we present this analysis for the entire Fourier domain, with noise sampled with norm $4$. For ImageNet, we focus our analysis on lower frequencies that are more visually salient add noise with norm $15.7$.%

Note that for \aug{Cutout} and \aug{Gaussian}, we chose larger patch sizes and $\sigma$s than those selected with the method in Section \ref{hyper-selection} in order to highlight the effect of these augmentations on sensitivity. Heatmaps of other models can be found in the Appendix (Figure \ref{full-filters}).

\subsection{Frequency-based analysis of models trained with Patch Gaussian}
\label{section-fourier-analysis}
We confirm findings by~\cite{yin2019fourier} that \aug{Gaussian} encourages the model to learn a low-pass filter of the inputs. Models trained with this augmentation, then, have low test error sensitivity at high frequencies, which could help robustness. However, valuable high-frequency information~\citep{brendel2019approximating} is being thrown out at low layers, which could explain the lower test accuracy.

We further find that \aug{Cutout} encourages the use of high-frequency information, which could help explain its improved generalization performance. Yet, it does not encourage lower test error sensitivity, which explains why it doesn't improve robustness either.

\begin{figure}[ht]
\begin{center}
  \centering
  \centerline{
      \begin{tabular}{rcccc|cc}
        & \multicolumn{2}{c}{Wide Resnet - CIFAR-10} & \multicolumn{2}{c|}{Resnet-50 - ImageNet} & Wide Resnet - CIFAR-10\\
        \cmidrule(r){2-3} \cmidrule(r){4-5} \cmidrule(r){6-6}
        & \linebreakcell{1st Layer\\Fourier\\Sensitivity} & \linebreakcell{Test Error\\Fourier\\Sensitivity} & \linebreakcell{Selected\\1st Layer\\Filters} & \linebreakcell{Test Error\\Fourier\\Sensitivity} & \multirow{6}{*}{\includegraphics[width=0.35\textwidth]{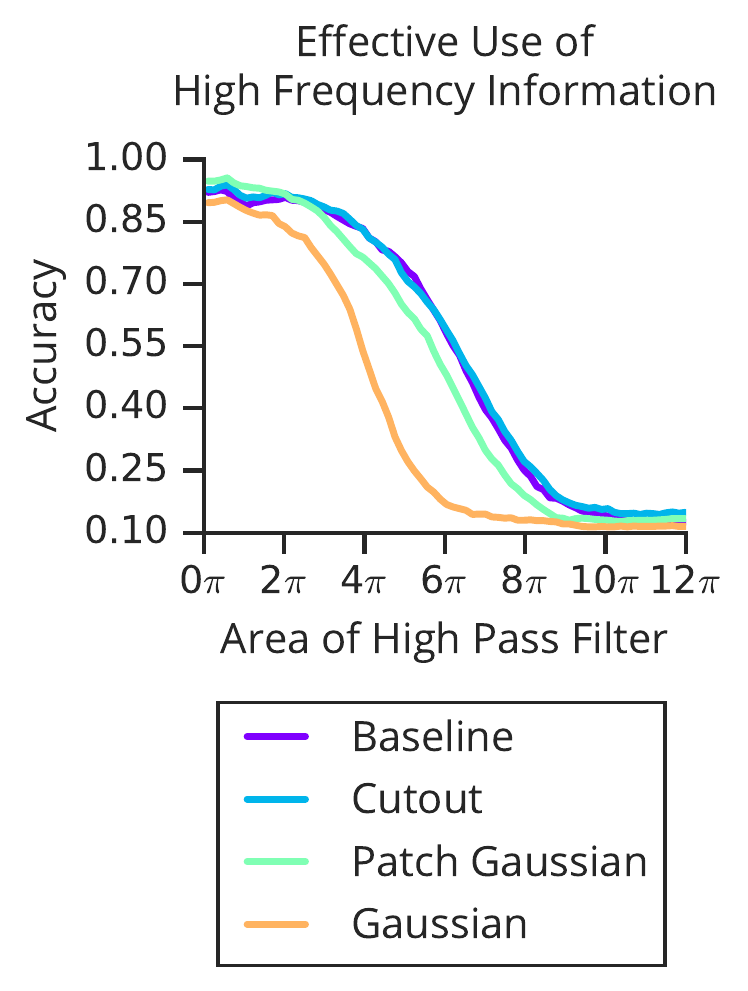}} & \raisebox{2.0\normalbaselineskip}[0pt][0pt]{\rotatebox[origin=c]{90}{\linebreakcell{\\}}}\\
        \raisebox{2.0\normalbaselineskip}[0pt][0pt]{\rotatebox[origin=c]{90}{\aug{Baseline}}} & \includegraphics[width=0.12\textwidth]{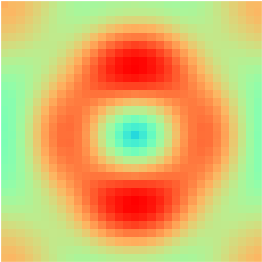} & \includegraphics[width=0.12\textwidth]{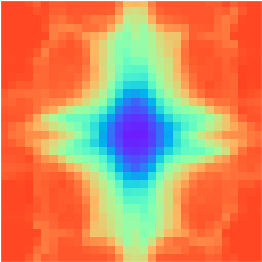} & 
            \raisebox{2.0\normalbaselineskip}[0pt][0pt]{
                \begin{tabular}{@{\hspace{0em}}c@{\hspace{0.35em}}c@{\hspace{0em}}}
                    \includegraphics[width=0.035\textwidth]{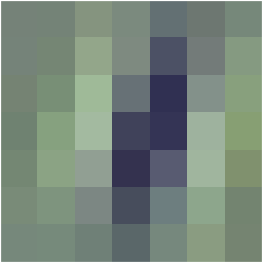} & \includegraphics[width=0.035\textwidth]{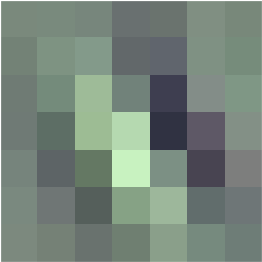}\\
                    \includegraphics[width=0.035\textwidth]{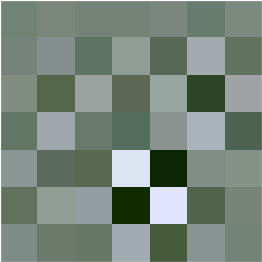} & \includegraphics[width=0.035\textwidth]{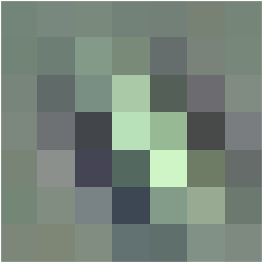}\\
                 \end{tabular}
             }
        & \includegraphics[width=0.12\textwidth]{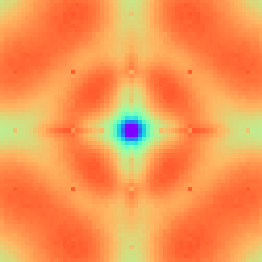} \\
        \raisebox{2.0\normalbaselineskip}[0pt][0pt]{\rotatebox[origin=c]{90}{\aug{Cutout}*}} & \includegraphics[width=0.12\textwidth]{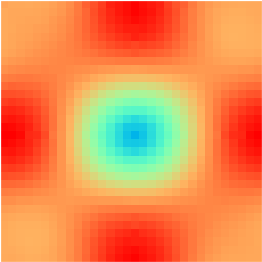} & \includegraphics[width=0.12\textwidth]{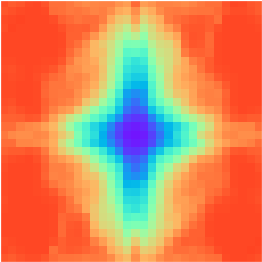} &
            \raisebox{2.0\normalbaselineskip}[0pt][0pt]{
                \begin{tabular}{@{\hspace{0em}}c@{\hspace{0.35em}}c@{\hspace{0em}}}
                    \includegraphics[width=0.035\textwidth]{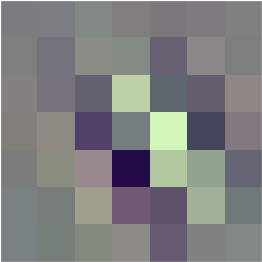} & \includegraphics[width=0.035\textwidth]{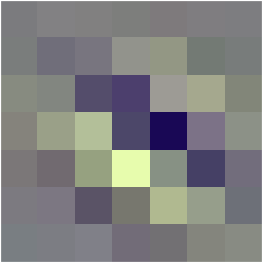}\\
                    \includegraphics[width=0.035\textwidth]{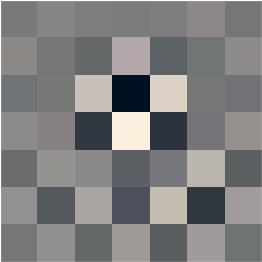} & \includegraphics[width=0.035\textwidth]{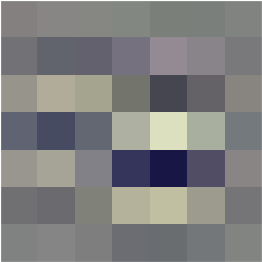}\\
                 \end{tabular}
             }
             & \includegraphics[width=0.12\textwidth]{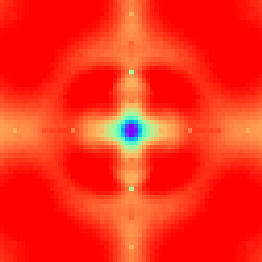}\\
        \raisebox{2.0\normalbaselineskip}[0pt][0pt]{\rotatebox[origin=c]{90}{\aug{Gaussian}*}} & \includegraphics[width=0.12\textwidth]{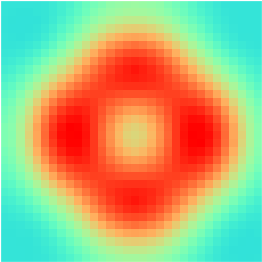} & \includegraphics[width=0.12\textwidth]{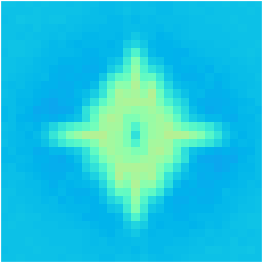} & 
            \raisebox{2.0\normalbaselineskip}[0pt][0pt]{
                \begin{tabular}{@{\hspace{0em}}c@{\hspace{0.35em}}c@{\hspace{0em}}}
                    \includegraphics[width=0.035\textwidth]{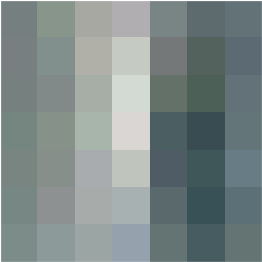} & \includegraphics[width=0.035\textwidth]{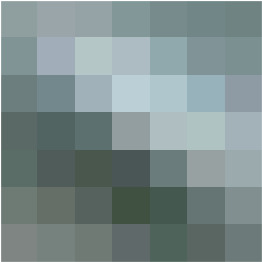}\\
                    \includegraphics[width=0.035\textwidth]{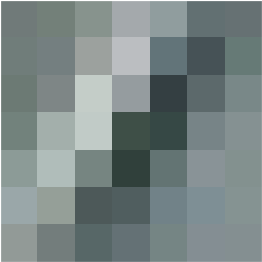} & \includegraphics[width=0.035\textwidth]{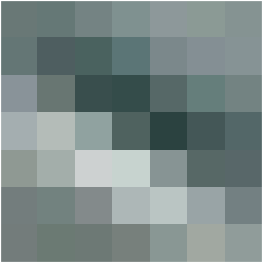}\\
                 \end{tabular}
             }
         & \includegraphics[width=0.12\textwidth]{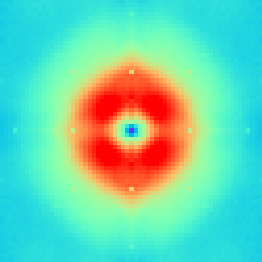}\\
        \raisebox{2.0\normalbaselineskip}[0pt][0pt]{\rotatebox[origin=c]{90}{\linebreakcell{\aug{Patch}\\\aug{Gaussian}}}} & \includegraphics[width=0.12\textwidth]{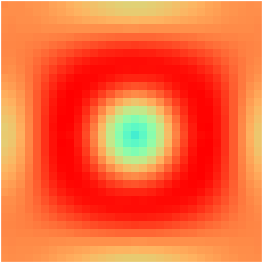} & \includegraphics[width=0.12\textwidth]{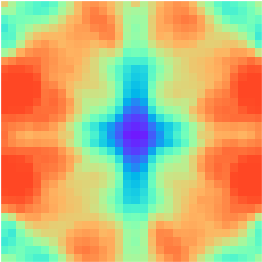} & 
            \raisebox{2.0\normalbaselineskip}[0pt][0pt]{
                 \begin{tabular}{@{\hspace{0em}}c@{\hspace{0.35em}}c@{\hspace{0em}}}
                    \includegraphics[width=0.035\textwidth]{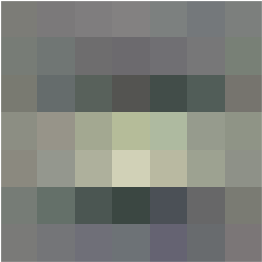} & \includegraphics[width=0.035\textwidth]{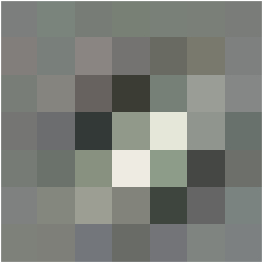}\\
                    \includegraphics[width=0.035\textwidth]{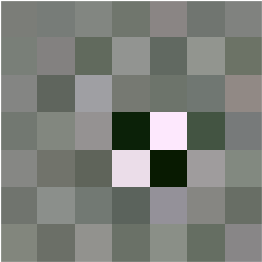} & \includegraphics[width=0.035\textwidth]{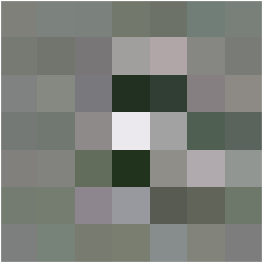}\\
                 \end{tabular}
            }
         & \includegraphics[width=0.12\textwidth]{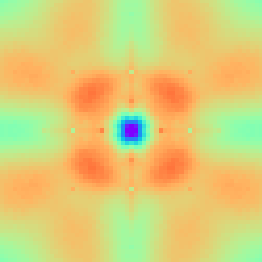} \\
        & \includegraphics[width=0.14\textwidth]{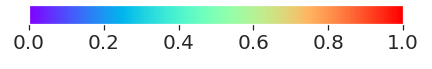} & \includegraphics[width=0.14\textwidth]{figures/fourier/colorbar.png} & & \includegraphics[width=0.14\textwidth]{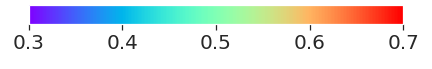}\\
      \end{tabular}
    }
\caption{
(left) Fourier analysis of various models, using method from \cite{yin2019fourier}.
Heatmaps depict model sensitivity to various sinusoidal gratings. 
\aug{Cutout} encourages the use of high frequencies in earlier layers, but its test error remains too sensitive to them. \aug{Gaussian} learns low-pass filtering of features, which increases robustness at later layers, but makes lower layers too invariant to high-frequency information (thus hurting accuracy). \aug{Patch Gaussian} allows high frequencies to be used in lower layers, and its test error remains relatively robust to them.
This can also be seen by the presence of high-frequency kernels in the first layer filters of the models (or lack thereof, in the case of \aug{Gaussian}).
(right) Indeed, \aug{Patch Gaussian} models match the performance of \aug{Cutout} and \aug{Baseline} when presented with only the high frequency information of images, whereas \aug{Gaussian} fails to effectively utilize this information (see Appendix Fig. \ref{highpass-deets} for experiment details).
This pattern of reduced sensitivity of predictions to high frequencies in the input occurs across all augmentation magnitudes, but here we use larger patch sizes and $\sigma$ of noise to highlight the differences in models indicated by *. See text for details.
}
\label{fourier}
\end{center}
\vspace{-0.2in}
\end{figure}

\aug{Patch Gaussian}, on the other hand, seems to allow high-frequency information through at lower layers, but still encourages relatively lower test error sensitivity at high frequencies. Indeed, when we measure accuracy on images filtered with a high-pass filter, we see that \aug{Patch Gaussian} models can maintain accuracy in a similar way to the baseline and to \aug{Cutout}, where \aug{Gaussian} fails to. See Figure \ref{fourier} for full results. 

Understanding the impact of data distributions and noise on representations has been well-studied in neuroscience \citep{barlow1961possible, simoncelli2001natural, karklin2011efficient}. The data augmentations that we propose here alter the distribution of inputs that the network sees, and thus are expected to alter the kinds of representations that are learned. Prior work on efficient coding \citep{karklin2011efficient} and autoencoders \citep{poole2014analyzing} has shown how filter properties change with noise in the unsupervised setting, resulting in lower-frequency filters with \aug{Gaussian}, as we observe in Fig.~\ref{fourier}. Consistent with prior work on natural image statistics \citep{torralba2003statistics}, we find that networks are least sensitive to low frequency noise where the spectral density is largest. Performance drops at higher frequencies when the amount of noise we add grows relative to the typical spectral density observed at these frequencies. In future work, we hope to better understand the relationship between naturally occuring properties of images and sensitivity, and investigate whether training with more naturalistic noise can yield similar gains in corruption robustness.

\newpage
\section{Conclusion}

In this work, we introduced a single data augmentation operation, \aug{Patch Gaussian}, which improves robustness to common corruptions without incurring a drop in clean accuracy. For models that are large relative to the dataset size (like ResNet-200 on ImageNet and all models on CIFAR-10), \aug{Patch Gaussian} improves clean accuracy and robustness concurrently. We showed that \aug{Patch Gaussian} achieves this by interpolating between two standard data augmentation operations \aug{Cutout} and \aug{Gaussian}.
We also demonstrate that \aug{Patch Gaussian} can be used in conjunction with other regularization and data augmentation strategies, and can also improve the performance of object detection models, indicating it is generally useful.
Finally, we analyzed the sensitivity to noise in different frequencies of models trained with \aug{Cutout} and \aug{Gaussian}, and showed that \aug{Patch Gaussian} combines their strengths without inheriting their weaknesses.

\subsubsection*{Acknowledgements}
We would like to thank Benjamin Caine, Trevor Gale, Keren Gu, Jonathon Shlens, Brandon Yang, Nic Ford, Samy Bengio, Alex Alemi, Matthew Streeter, Robert Geirhos, Alex Berardino, Jon Barron, Barret Zoph, and the Google Brain and Google AI Residency teams.

\medskip

{\small
\bibliography{bibliography}

\begin{thebibliography}{55}
\providecommand{\natexlab}[1]{#1}
\providecommand{\url}[1]{\texttt{#1}}
\expandafter\ifx\csname urlstyle\endcsname\relax
  \providecommand{\doi}[1]{doi: #1}\else
  \providecommand{\doi}{doi: \begingroup \urlstyle{rm}\Url}\fi

\bibitem[Asano et~al.(2019)Asano, Rupprecht, and Vedaldi]{1904.13132}
Asano, Y.~M., Rupprecht, C., and Vedaldi, A.
\newblock Surprising effectiveness of few-image unsupervised feature learning,
  2019.

\bibitem[Azulay \& Weiss(2018)Azulay and Weiss]{azulay2018deep}
Azulay, A. and Weiss, Y.
\newblock Why do deep convolutional networks generalize so poorly to small
  image transformations?
\newblock \emph{arXiv preprint arXiv:1805.12177}, 2018.

\bibitem[Barlow et~al.(1961)]{barlow1961possible}
Barlow, H.~B. et~al.
\newblock Possible principles underlying the transformation of sensory
  messages.
\newblock \emph{Sensory communication}, 1:\penalty0 217--234, 1961.

\bibitem[Brendel \& Bethge(2019)Brendel and Bethge]{brendel2019approximating}
Brendel, W. and Bethge, M.
\newblock Approximating cnns with bag-of-local-features models works
  surprisingly well on imagenet.
\newblock \emph{arXiv preprint arXiv:1904.00760}, 2019.

\bibitem[Cubuk et~al.(2017)Cubuk, Zoph, Schoenholz, and
  Le]{cubuk2017intriguing}
Cubuk, E.~D., Zoph, B., Schoenholz, S.~S., and Le, Q.~V.
\newblock Intriguing properties of adversarial examples.
\newblock \emph{arXiv preprint arXiv:1711.02846}, 2017.

\bibitem[Cubuk et~al.(2018)Cubuk, Zoph, Mane, Vasudevan, and
  Le]{cubuk2018autoaugment}
Cubuk, E.~D., Zoph, B., Mane, D., Vasudevan, V., and Le, Q.~V.
\newblock Autoaugment: Learning augmentation policies from data.
\newblock \emph{arXiv preprint arXiv:1805.09501}, 2018.

\bibitem[Dao et~al.(2018)Dao, Gu, Ratner, Smith, De~Sa, and
  R{\'e}]{dao2018kernel}
Dao, T., Gu, A., Ratner, A.~J., Smith, V., De~Sa, C., and R{\'e}, C.
\newblock A kernel theory of modern data augmentation.
\newblock \emph{arXiv preprint arXiv:1803.06084}, 2018.

\bibitem[Deng et~al.(2009)Deng, Dong, Socher, Li, Li, and
  Fei-Fei]{imagenet2009}
Deng, J., Dong, W., Socher, R., Li, L.-J., Li, K., and Fei-Fei, L.
\newblock Imagenet: A large-scale hierarchical image database.
\newblock In \emph{Proceedings of IEEE Conference on Computer Vision and
  Pattern Recognition (CVPR)}, 2009.

\bibitem[DeVries \& Taylor(2017)DeVries and Taylor]{devries2017improved}
DeVries, T. and Taylor, G.~W.
\newblock Improved regularization of convolutional neural networks with cutout.
\newblock \emph{arXiv preprint arXiv:1708.04552}, 2017.

\bibitem[Dodge \& Karam(2017)Dodge and Karam]{dodge2017study}
Dodge, S. and Karam, L.
\newblock A study and comparison of human and deep learning recognition
  performance under visual distortions.
\newblock In \emph{2017 26th international conference on computer communication
  and networks (ICCCN)}, pp.\  1--7. IEEE, 2017.

\bibitem[Doersch et~al.(2015)Doersch, Gupta, and
  Efros]{doersch2015unsupervised}
Doersch, C., Gupta, A., and Efros, A.~A.
\newblock Unsupervised visual representation learning by context prediction.
\newblock In \emph{Proceedings of the IEEE International Conference on Computer
  Vision}, pp.\  1422--1430, 2015.

\bibitem[{ECMA International}(2009)]{jpeg}
{ECMA International}.
\newblock \emph{{JPEG Interchange Format (JFIF)}}.
\newblock 2009.

\bibitem[Ford et~al.(2019)Ford, Gilmer, Carlini, and
  Cubuk]{ford2019adversarial}
Ford, N., Gilmer, J., Carlini, N., and Cubuk, D.
\newblock Adversarial examples are a natural consequence of test error in
  noise.
\newblock \emph{arXiv preprint arXiv:1901.10513}, 2019.

\bibitem[Gastaldi(2017)]{gastaldi2017shake}
Gastaldi, X.
\newblock Shake-shake regularization.
\newblock \emph{arXiv preprint arXiv:1705.07485}, 2017.

\bibitem[Geirhos et~al.(2018{\natexlab{a}})Geirhos, Rubisch, Michaelis, Bethge,
  Wichmann, and Brendel]{geirhos2018imagenet}
Geirhos, R., Rubisch, P., Michaelis, C., Bethge, M., Wichmann, F.~A., and
  Brendel, W.
\newblock Imagenet-trained cnns are biased towards texture; increasing shape
  bias improves accuracy and robustness.
\newblock \emph{arXiv preprint arXiv:1811.12231}, 2018{\natexlab{a}}.

\bibitem[Geirhos et~al.(2018{\natexlab{b}})Geirhos, Temme, Rauber, Sch{\"u}tt,
  Bethge, and Wichmann]{geirhos2018generalisation}
Geirhos, R., Temme, C.~R., Rauber, J., Sch{\"u}tt, H.~H., Bethge, M., and
  Wichmann, F.~A.
\newblock Generalisation in humans and deep neural networks.
\newblock In \emph{Advances in Neural Information Processing Systems}, pp.\
  7538--7550, 2018{\natexlab{b}}.

\bibitem[Ghiasi et~al.(2018)Ghiasi, Lin, and Le]{ghiasi2018dropblock}
Ghiasi, G., Lin, T.-Y., and Le, Q.~V.
\newblock Dropblock: A regularization method for convolutional networks.
\newblock In \emph{Advances in Neural Information Processing Systems}, pp.\
  10727--10737, 2018.

\bibitem[Gilmer et~al.(2018)Gilmer, Adams, Goodfellow, Andersen, and
  Dahl]{gilmer2018motivating}
Gilmer, J., Adams, R.~P., Goodfellow, I., Andersen, D., and Dahl, G.~E.
\newblock Motivating the rules of the game for adversarial example research.
\newblock \emph{arXiv preprint arXiv:1807.06732}, 2018.

\bibitem[Goodfellow et~al.(2014)Goodfellow, Shlens, and
  Szegedy]{goodfellow2014explaining}
Goodfellow, I.~J., Shlens, J., and Szegedy, C.
\newblock Explaining and harnessing adversarial examples.
\newblock \emph{arXiv preprint arXiv:1412.6572}, 2014.

\bibitem[Grandvalet \& Canu(1997)Grandvalet and Canu]{grandvalet1997noise}
Grandvalet, Y. and Canu, S.
\newblock Noise injection for inputs relevance determination.
\newblock \emph{Advances in intelligent systems}, 41:\penalty0 378, 1997.

\bibitem[Gu et~al.(2019)Gu, Yang, Ngiam, Le, and Shlens]{gu2019using}
Gu, K., Yang, B., Ngiam, J., Le, Q., and Shlens, J.
\newblock Using videos to evaluate image model robustness.
\newblock \emph{arXiv preprint arXiv:1904.10076}, 2019.

\bibitem[Han et~al.(2017)Han, Kim, and Kim]{han2017deep}
Han, D., Kim, J., and Kim, J.
\newblock Deep pyramidal residual networks.
\newblock In \emph{Proceedings of IEEE Conference on Computer Vision and
  Pattern Recognition (CVPR)}, pp.\  6307--6315. IEEE, 2017.

\bibitem[He et~al.(2016)He, Zhang, Ren, and Sun]{he2016deep}
He, K., Zhang, X., Ren, S., and Sun, J.
\newblock Deep residual learning for image recognition.
\newblock In \emph{Proceedings of the IEEE Conference on Computer Vision and
  Pattern Recognition (CVPR)}, pp.\  770--778, 2016.

\bibitem[Hendrycks \& Dietterich(2018)Hendrycks and
  Dietterich]{hendrycks2018benchmarking}
Hendrycks, D. and Dietterich, T.~G.
\newblock Benchmarking neural network robustness to common corruptions and
  surface variations.
\newblock \emph{arXiv preprint arXiv:1807.01697}, 2018.

\bibitem[Hendrycks et~al.(2019)Hendrycks, Lee, and Mazeika]{hendrycks2019using}
Hendrycks, D., Lee, K., and Mazeika, M.
\newblock Using pre-training can improve model robustness and uncertainty.
\newblock \emph{arXiv preprint arXiv:1901.09960}, 2019.

\bibitem[Hu et~al.(2017)Hu, Shen, and Sun]{hu2017squeeze}
Hu, J., Shen, L., and Sun, G.
\newblock Squeeze-and-excitation networks.
\newblock \emph{arXiv preprint arXiv:1709.01507}, 2017.

\bibitem[Huang et~al.(2018)Huang, Cheng, Chen, Lee, Ngiam, Le, and
  Chen]{huang2018gpipe}
Huang, Y., Cheng, Y., Chen, D., Lee, H., Ngiam, J., Le, Q.~V., and Chen, Z.
\newblock Gpipe: Efficient training of giant neural networks using pipeline
  parallelism.
\newblock \emph{arXiv preprint arXiv:1811.06965}, 2018.

\bibitem[Ilyas et~al.(2019)Ilyas, Santurkar, Tsipras, Engstrom, Tran, and
  Madry]{ilyas2019adversarial}
Ilyas, A., Santurkar, S., Tsipras, D., Engstrom, L., Tran, B., and Madry, A.
\newblock Adversarial examples are not bugs, they are features.
\newblock \emph{arXiv preprint arXiv:1905.02175}, 2019.

\bibitem[Jacobsen et~al.(2018)Jacobsen, Behrmann, Zemel, and
  Bethge]{jacobsen2018excessive}
Jacobsen, J.-H., Behrmann, J., Zemel, R., and Bethge, M.
\newblock Excessive invariance causes adversarial vulnerability.
\newblock \emph{arXiv preprint arXiv:1811.00401}, 2018.

\bibitem[Karklin \& Simoncelli(2011)Karklin and
  Simoncelli]{karklin2011efficient}
Karklin, Y. and Simoncelli, E.~P.
\newblock Efficient coding of natural images with a population of noisy
  linear-nonlinear neurons.
\newblock In \emph{Advances in neural information processing systems}, pp.\
  999--1007, 2011.

\bibitem[Karpathy(2011)]{karpathy2011lessons}
Karpathy, A.
\newblock Lessons learned from manually classifying cifar-10.
\newblock \emph{Published online at http://karpathy. github.
  io/2011/04/27/manually-classifying-cifar10}, 2011.

\bibitem[Krizhevsky \& Hinton(2009)Krizhevsky and
  Hinton]{krizhevsky2009learning}
Krizhevsky, A. and Hinton, G.
\newblock Learning multiple layers of features from tiny images.
\newblock Technical report, Citeseer, 2009.

\bibitem[Krizhevsky et~al.(2012)Krizhevsky, Sutskever, and
  Hinton]{krizhevsky2012imagenet}
Krizhevsky, A., Sutskever, I., and Hinton, G.~E.
\newblock Imagenet classification with deep convolutional neural networks.
\newblock In \emph{Advances in Neural Information Processing Systems}, 2012.

\bibitem[Lin et~al.(2014)Lin, Maire, Belongie, Hays, Perona, Ramanan,
  Doll{\'a}r, and Zitnick]{lin2014microsoft}
Lin, T.-Y., Maire, M., Belongie, S., Hays, J., Perona, P., Ramanan, D.,
  Doll{\'a}r, P., and Zitnick, C.~L.
\newblock Microsoft coco: Common objects in context.
\newblock In \emph{European conference on computer vision}, pp.\  740--755.
  Springer, 2014.

\bibitem[Lin et~al.(2017)Lin, Goyal, Girshick, He, and
  Doll{\'a}r]{lin2017focal}
Lin, T.-Y., Goyal, P., Girshick, R., He, K., and Doll{\'a}r, P.
\newblock Focal loss for dense object detection.
\newblock In \emph{Proceedings of the IEEE international conference on computer
  vision}, pp.\  2980--2988, 2017.

\bibitem[Liu et~al.(2018)Liu, Simonyan, and Yang]{liu2018darts}
Liu, H., Simonyan, K., and Yang, Y.
\newblock Darts: Differentiable architecture search.
\newblock \emph{arXiv preprint arXiv:1806.09055}, 2018.

\bibitem[Madry et~al.(2017)Madry, Makelov, Schmidt, Tsipras, and
  Vladu]{madry2017towards}
Madry, A., Makelov, A., Schmidt, L., Tsipras, D., and Vladu, A.
\newblock Towards deep learning models resistant to adversarial attacks.
\newblock \emph{arXiv preprint arXiv:1706.06083}, 2017.

\bibitem[Poole et~al.(2014)Poole, Sohl-Dickstein, and
  Ganguli]{poole2014analyzing}
Poole, B., Sohl-Dickstein, J., and Ganguli, S.
\newblock Analyzing noise in autoencoders and deep networks, 2014.

\bibitem[Recht et~al.(2018)Recht, Roelofs, Schmidt, and
  Shankar]{recht2018cifar}
Recht, B., Roelofs, R., Schmidt, L., and Shankar, V.
\newblock Do cifar-10 classifiers generalize to cifar-10?
\newblock \emph{arXiv preprint arXiv:1806.00451}, 2018.

\bibitem[Recht et~al.(2019)Recht, Roelofs, Schmidt, and
  Shankar]{recht2019imagenet}
Recht, B., Roelofs, R., Schmidt, L., and Shankar, V.
\newblock Do imagenet classifiers generalize to imagenet?
\newblock \emph{arXiv preprint arXiv:1902.10811}, 2019.

\bibitem[Rosenfeld et~al.(2018)Rosenfeld, Zemel, and
  Tsotsos]{rosenfeld2018elephant}
Rosenfeld, A., Zemel, R., and Tsotsos, J.~K.
\newblock The elephant in the room.
\newblock \emph{arXiv preprint arXiv:1808.03305}, 2018.

\bibitem[Simoncelli \& Olshausen(2001)Simoncelli and
  Olshausen]{simoncelli2001natural}
Simoncelli, E.~P. and Olshausen, B.~A.
\newblock Natural image statistics and neural representation.
\newblock \emph{Annual review of neuroscience}, 24\penalty0 (1):\penalty0
  1193--1216, 2001.

\bibitem[Simonyan \& Zisserman(2015)Simonyan and Zisserman]{simonyan2014very}
Simonyan, K. and Zisserman, A.
\newblock Very deep convolutional networks for large-scale image recognition.
\newblock \emph{Advances in Neural Information Processing Systems}, 2015.

\bibitem[Szegedy et~al.(2015)Szegedy, Liu, Jia, Sermanet, Reed, Anguelov,
  Erhan, Vanhoucke, Rabinovich, et~al.]{szegedy2015going}
Szegedy, C., Liu, W., Jia, Y., Sermanet, P., Reed, S., Anguelov, D., Erhan, D.,
  Vanhoucke, V., Rabinovich, A., et~al.
\newblock Going deeper with convolutions.
\newblock In \emph{Proceedings of the IEEE Conference on Computer Vision and
  Pattern Recognition (CVPR)}, 2015.

\bibitem[Szegedy et~al.(2017)Szegedy, Ioffe, Vanhoucke, and
  Alemi]{szegedy2017inception}
Szegedy, C., Ioffe, S., Vanhoucke, V., and Alemi, A.~A.
\newblock Inception-v4, inception-resnet and the impact of residual connections
  on learning.
\newblock In \emph{AAAI}, 2017.

\bibitem[Torralba \& Oliva(2003)Torralba and Oliva]{torralba2003statistics}
Torralba, A. and Oliva, A.
\newblock Statistics of natural image categories.
\newblock \emph{Network: computation in neural systems}, 14\penalty0
  (3):\penalty0 391--412, 2003.

\bibitem[Tsipras et~al.(2018)Tsipras, Santurkar, Engstrom, Turner, and
  Madry]{tsipras2018robustness}
Tsipras, D., Santurkar, S., Engstrom, L., Turner, A., and Madry, A.
\newblock Robustness may be at odds with accuracy.
\newblock \emph{stat}, 1050:\penalty0 11, 2018.

\bibitem[Wang et~al.(2019)Wang, He, Lipton, and Xing]{wang2019learning}
Wang, H., He, Z., Lipton, Z.~C., and Xing, E.~P.
\newblock Learning robust representations by projecting superficial statistics
  out.
\newblock \emph{arXiv preprint arXiv:1903.06256}, 2019.

\bibitem[Xie et~al.(2019)Xie, Dai, Hovy, Luong, and Le]{1904.12848}
Xie, Q., Dai, Z., Hovy, E., Luong, M.-T., and Le, Q.~V.
\newblock Unsupervised data augmentation, 2019.

\bibitem[Yin et~al.(2019)Yin, Gontijo~Lopes, Shlens, Cubuk, and
  Gilmer]{yin2019fourier}
Yin, D., Gontijo~Lopes, R., Shlens, J., Cubuk, E.~D., and Gilmer, J.
\newblock A fourier perspective on model robustness in computer vision.
\newblock \emph{ICML Workshop on Uncertainty and Robustness in Deep Learning},
  2019.

\bibitem[Zagoruyko \& Komodakis(2016)Zagoruyko and
  Komodakis]{zagoruyko2016wide}
Zagoruyko, S. and Komodakis, N.
\newblock Wide residual networks.
\newblock \emph{arXiv preprint arXiv:1605.07146}, 2016.

\bibitem[Zhang et~al.(2017)Zhang, Cisse, Dauphin, and
  Lopez-Paz]{zhang2017mixup}
Zhang, H., Cisse, M., Dauphin, Y.~N., and Lopez-Paz, D.
\newblock mixup: Beyond empirical risk minimization.
\newblock \emph{arXiv preprint arXiv:1710.09412}, 2017.

\bibitem[Zhong et~al.(2017)Zhong, Zheng, Kang, Li, and Yang]{zhong2017random}
Zhong, Z., Zheng, L., Kang, G., Li, S., and Yang, Y.
\newblock Random erasing data augmentation.
\newblock \emph{arXiv preprint arXiv:1708.04896}, 2017.

\bibitem[Zoph \& Le(2017)Zoph and Le]{zoph2016neural}
Zoph, B. and Le, Q.~V.
\newblock Neural architecture search with reinforcement learning.
\newblock In \emph{International Conference on Learning Representations}, 2017.

\bibitem[Zoph et~al.(2017)Zoph, Vasudevan, Shlens, and Le]{zoph2017learning}
Zoph, B., Vasudevan, V., Shlens, J., and Le, Q.~V.
\newblock Learning transferable architectures for scalable image recognition.
\newblock In \emph{Proceedings of IEEE Conference on Computer Vision and
  Pattern Recognition}, 2017.

\end{thebibliography}
\bibliographystyle{icml2019}}

\newpage
\section*{Appendix}
\begin{figure*}[h]
\begin{center}
\begin{minipage}[c]{0.48\textwidth}
\centerline{\includegraphics[width=\textwidth]{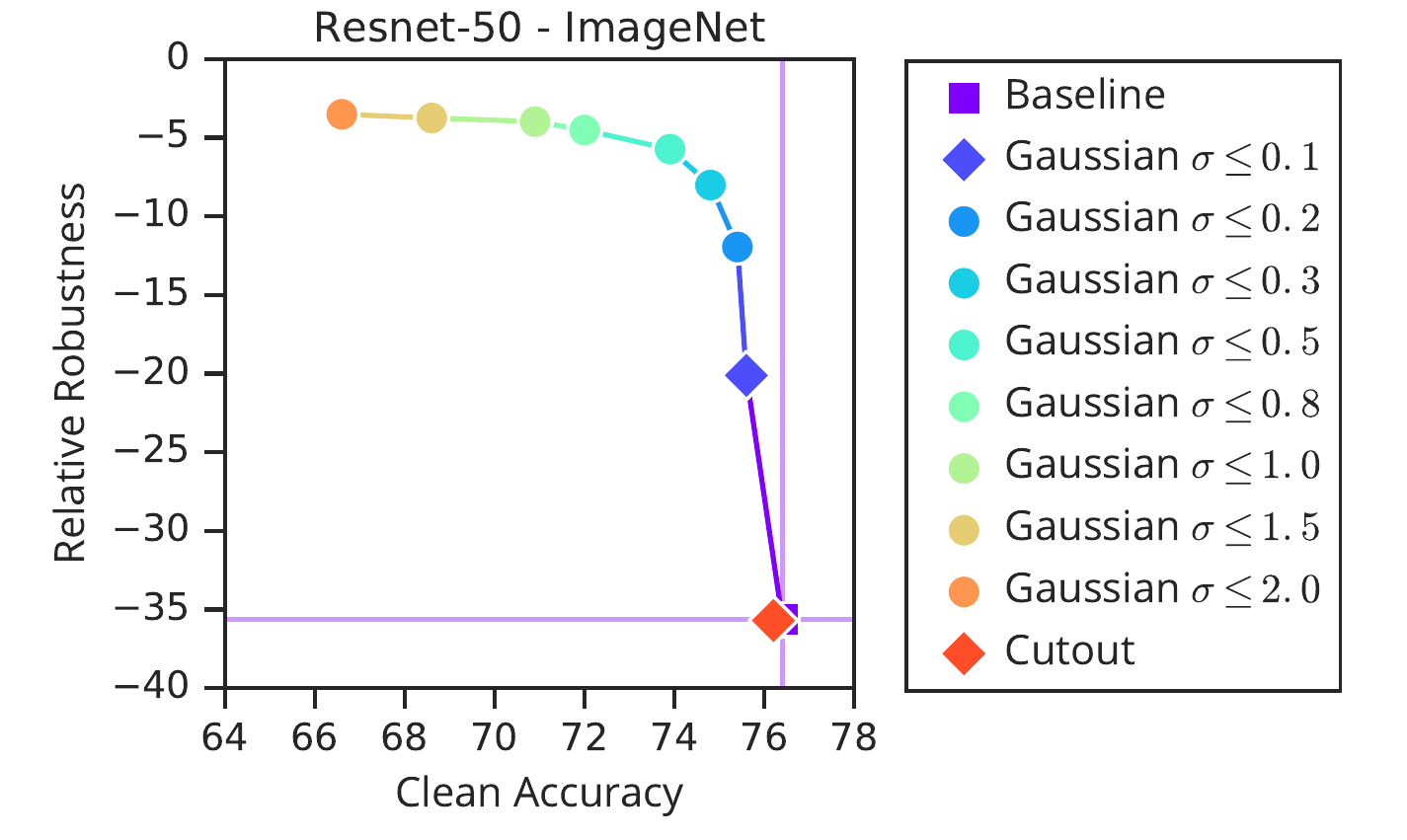}}
\end{minipage}%
\caption{Accuracy/robustness trade-off observed for \aug{Cutout} and \aug{Gaussian} on Resnet-50 models. See Figure \ref{cutout-vs-fullgauss} for details.}
\label{cutout-gaussian-tradeoff-rn50}
\end{center}
\end{figure*}%

\begin{figure*}[h]
\begin{center}
\centerline{
\begin{tabular}{cc}
        Wide-Resnet - CIFAR-10 & Resnet-50 - ImageNet  \\
        \cmidrule(r){1-1} \cmidrule(r){2-2}
        \includegraphics[width=0.48\textwidth]{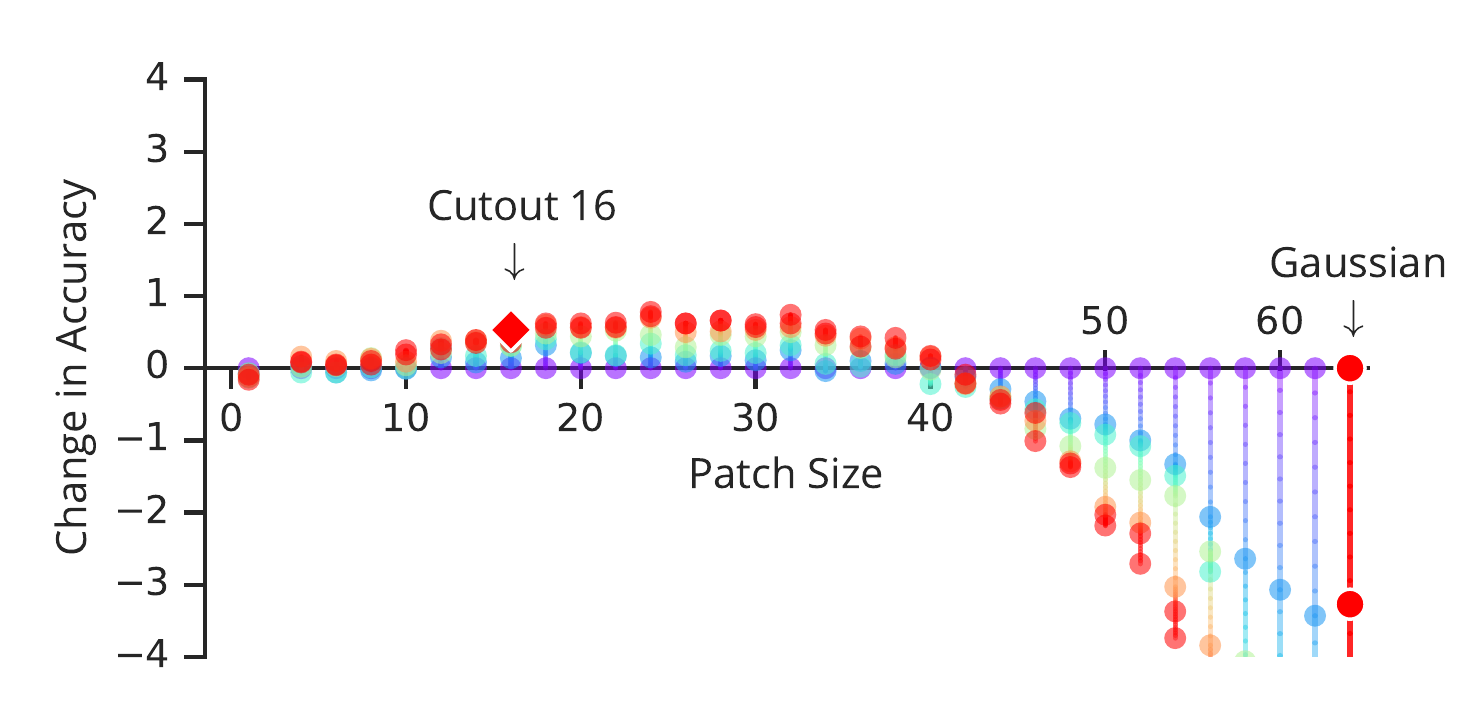} & \includegraphics[width=0.48\textwidth]{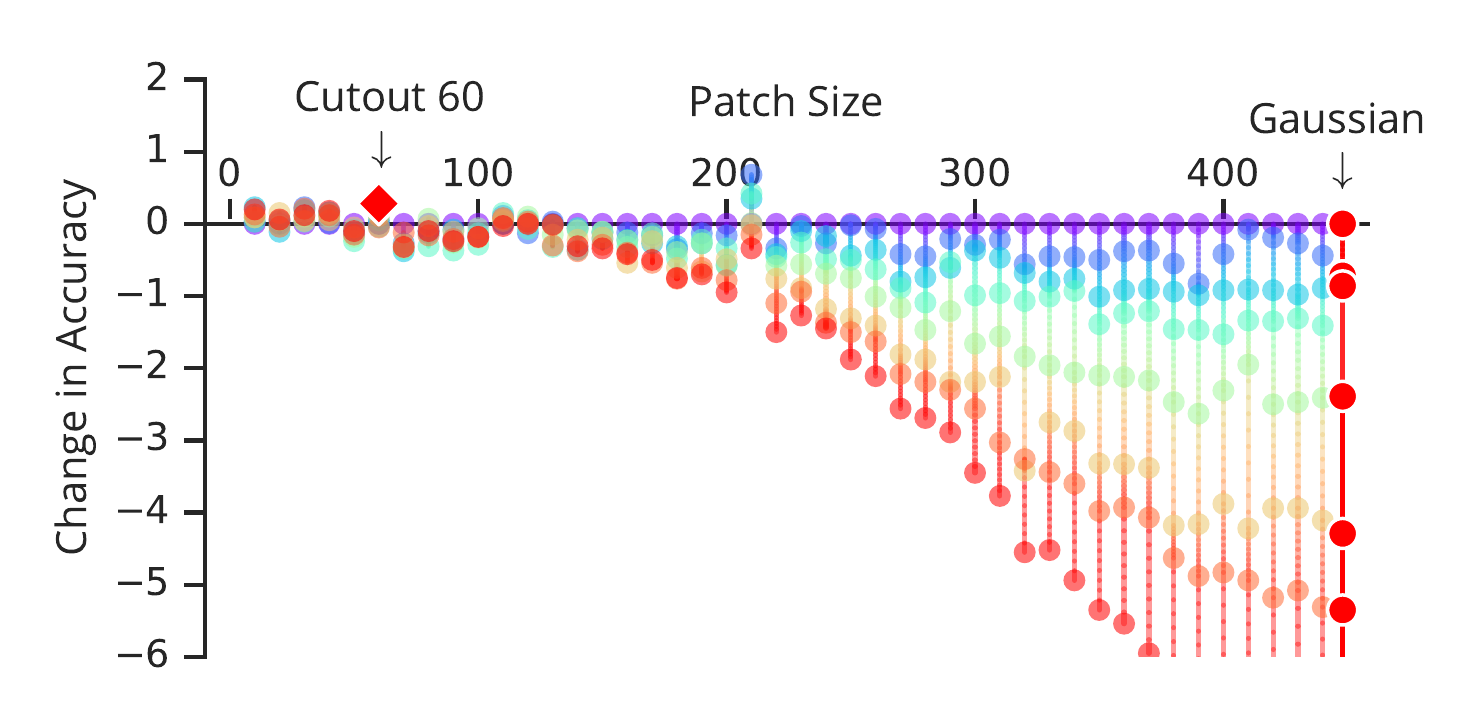}\\
\end{tabular}
}
\caption{
\aug{Patch Gaussian} hyper-parameter sweep for Wide-Resnet on CIFAR-10 (left) and RN50 on Imagenet (right).
\aug{Patch Gaussian} approaches \aug{Gaussian} with increasing patch size and \aug{Cutout} with increasing $\sigma$. Each dot is a model trained with different hyper parameters. 
Colors indicate different $\sigma$.
}
\label{patch-gauss-sweeps}
\end{center}
\end{figure*}%

\begin{figure}[ht!]
\begin{center}
    \begin{minipage}[c]{0.48\textwidth}
    \centerline{\includegraphics[width=\textwidth]{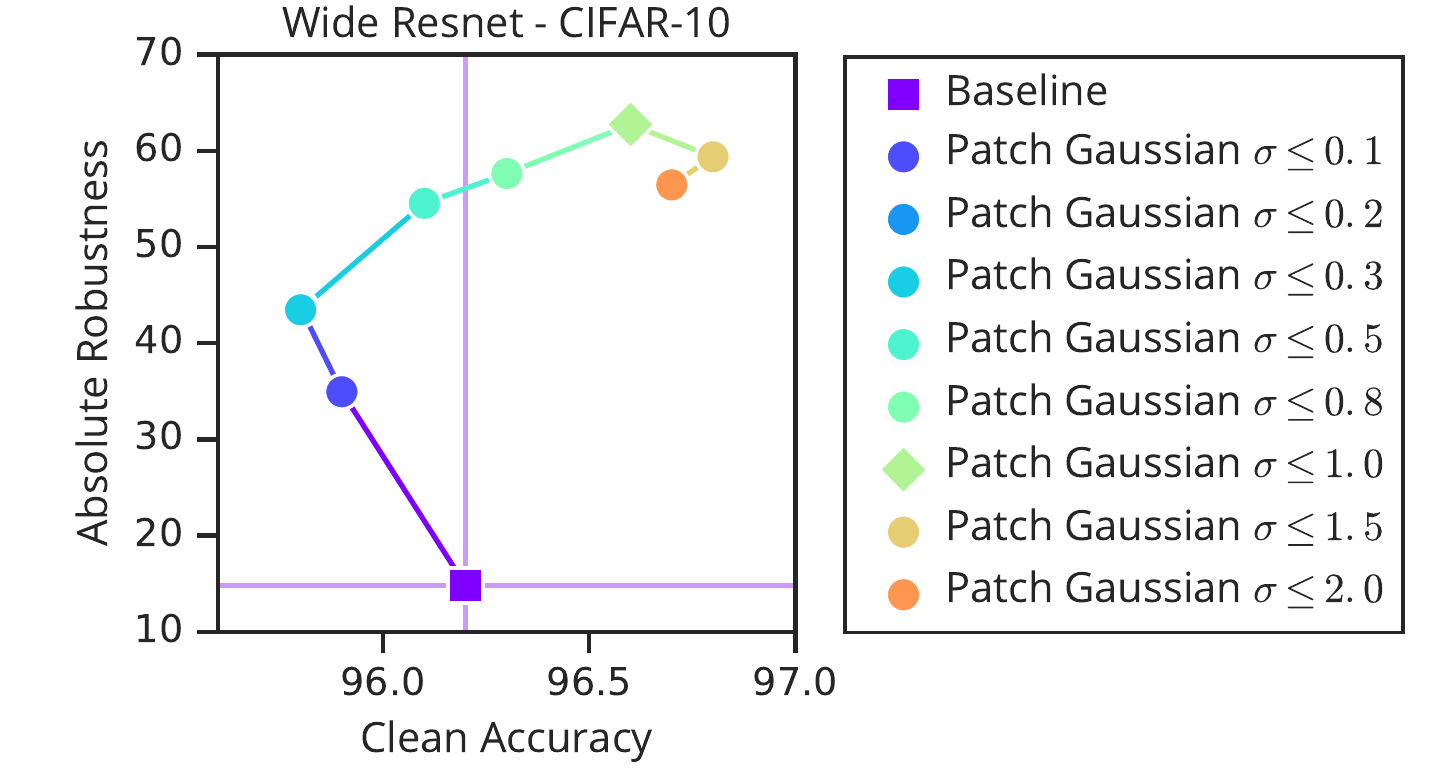}}
    \end{minipage}%
    \begin{minipage}[c]{0.48\textwidth}
    \centerline{\includegraphics[width=\textwidth]{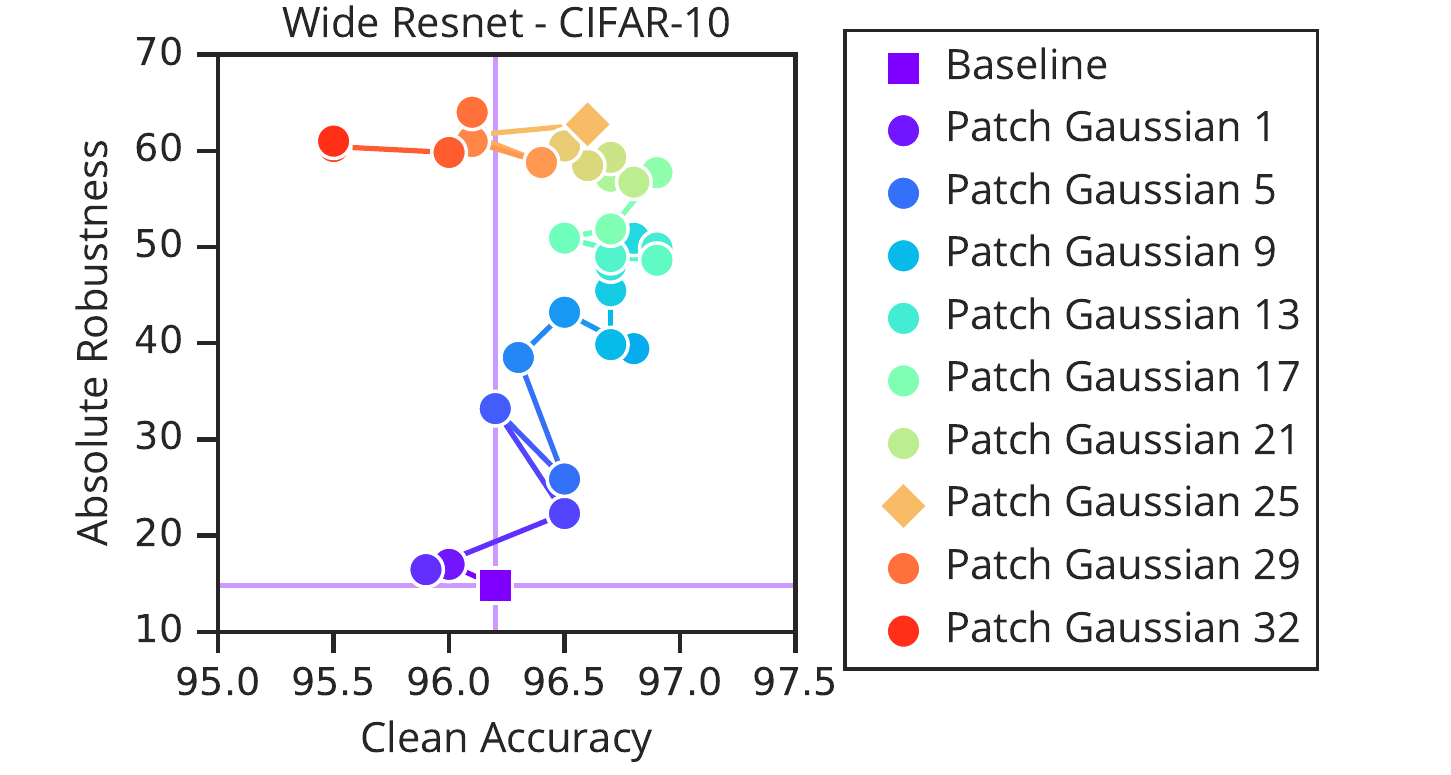}}
    \end{minipage}
    \begin{minipage}[c]{0.48\textwidth}
    \centerline{\includegraphics[width=\textwidth]{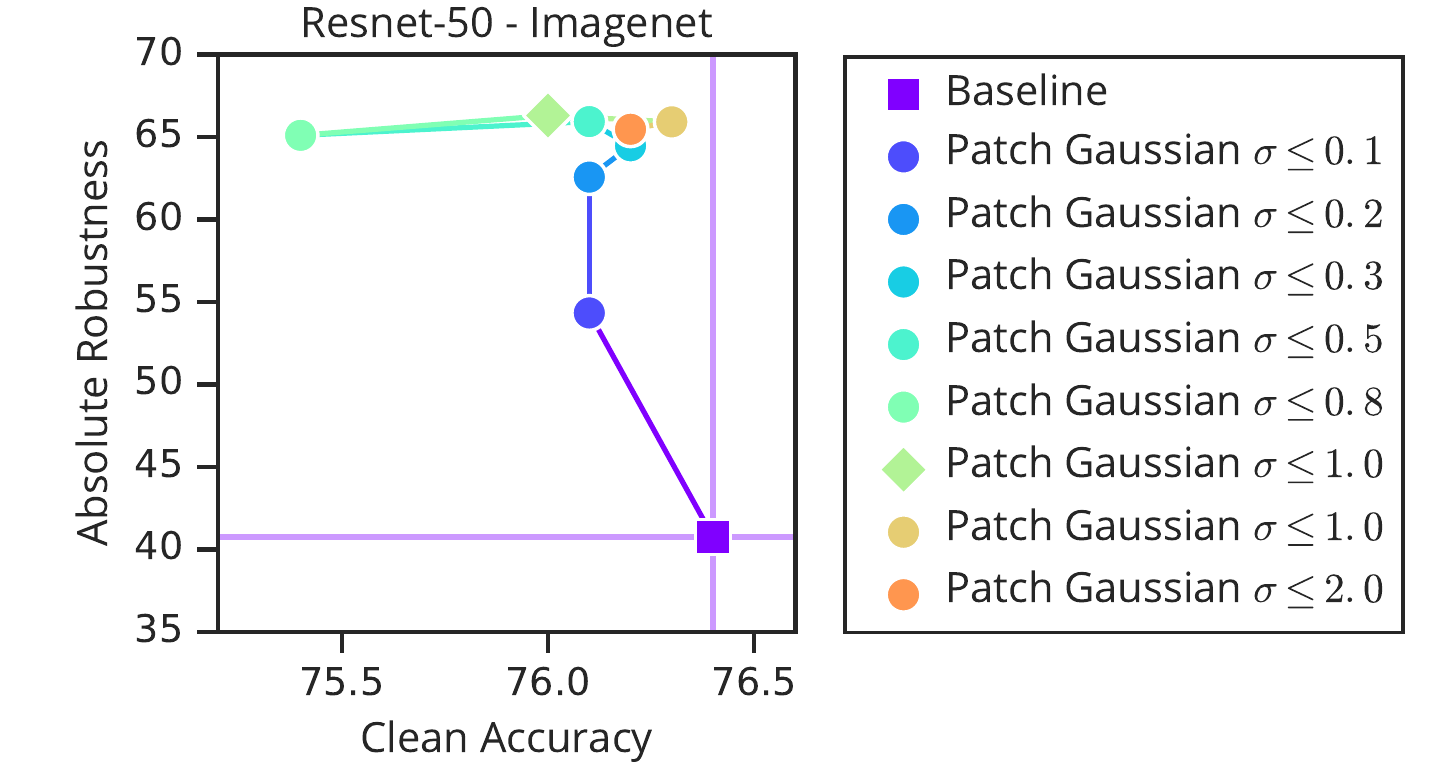}}
    \end{minipage}%
    \begin{minipage}[c]{0.48\textwidth}
    \centerline{\includegraphics[width=\textwidth]{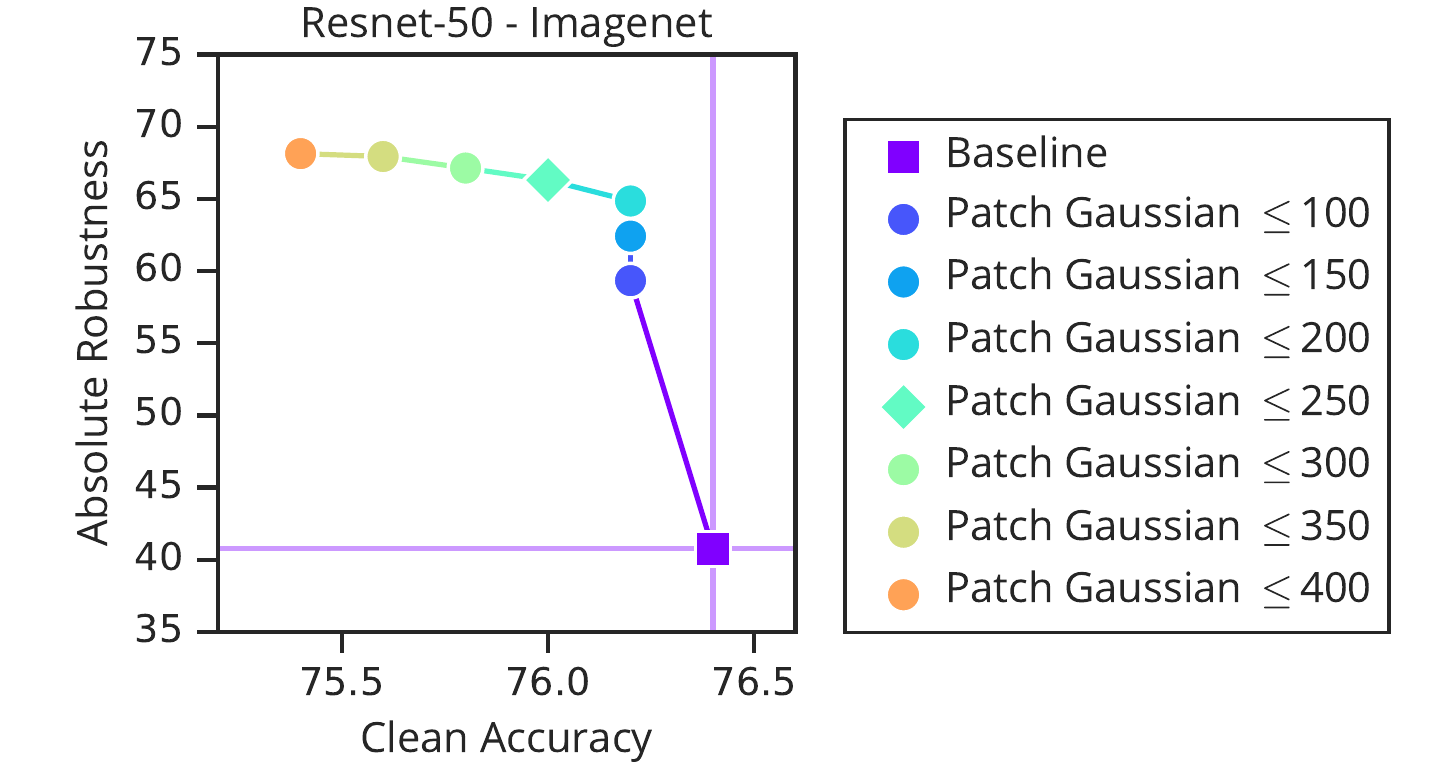}}
    \end{minipage}%
\caption{Overcoming the accuracy/robustness trade-off with \aug{Patch Gaussian} for models trained on CIFAR-10 (top row) and Resnet-50 (bottom row). See figure \ref{patch-gauss-overcoming} for details.
}
\label{patch-gauss-ablation-other-models}
\end{center}
\end{figure}%

\begin{figure}
\begin{verbatim}
def _get_patch_mask(patch_size):
  # randomly sample location in the image
  x = tf.random.uniform([], minval=0, maxval=224, dtype=tf.int32)
  y = tf.random.uniform([], minval=0, maxval=224, dtype=tf.int32)
  x, y = tf.cast(x, tf.float32), tf.cast(y, tf.float32)

  # compute where the patch will start and end
  startx, starty = x - tf.floor(patch_size/2), y - tf.floor(patch_size/2)
  endx, endy = x + tf.ceil(patch_size/2), y + tf.ceil(patch_size/2)
  startx, starty = tf.maximum(startx, 0), tf.maximum(starty, 0)
  endx, endy = tf.minimum(endx, 224), tf.minimum(endy, 224)

  # now let's convert these into how much we need to pad the patch
  lower_pad, upper_pad = 224 - endy, starty
  left_pad, right_pad = startx, 224 - endx
  padding_dims = [[upper_pad, lower_pad], [left_pad, right_pad]]

  # create mask
  mask = tf.pad(tf.zeros([endy - starty, endx - startx]),
                padding_dims, constant_values=1)
  mask = tf.expand_dims(mask, -1)
  mask = tf.tile(mask, [1, 1, 3])
  return tf.equal(mask, 0)


def patch_gaussian(image, patch_size, max_scale, sample_up_to):
  """Returns image with Patch Gaussian applied."""

  if sample_up_to:
    patch_size = tf.random.uniform([], 1, patch_size, tf.int32)
    # otherwise, patch_size is fixed.
    
  # make image (which is [0, 255]) be [0, 1]
  image = image / 255.0

  # uniformly sample scale from 0 to given scale
  scale = max_scale * tf.random.uniform([], minval=0, maxval=1)

  # apply gaussian to copy of image. Will be used to replace patch in image
  gaussian = tf.random.normal(stddev=scale, shape=image.shape)
  image_plus_gaussian = tf.clip_by_value(image + gaussian, 0, 1)

  # create mask and apply patch
  image = tf.where(_get_patch_mask(patch_size),
                   image_plus_gaussian, image)

  # scale back to [0, 255]
  return image * 255
\end{verbatim}
\caption{TensorFlow implementation of \aug{Patch Gaussian}}
\label{implementation}
\end{figure}

\begin{table}
  \caption{Augmentation hyper-parameters selected with the method in Section \ref{hyper-selection} for each model/dataset. *Indicates manually-chosen stronger hyper-parameters, used to highlight the effect of the augmentation on the models. ``$\leq$'' indicates that the value is uniformly sampled up to the given maximum value.}
  \label{hyper-table}
  \centering
  \centerline{
  \begin{tabular}{llclccc}
    \toprule
        & & Z & Augmentation & $W$ & $\sigma$ & Other \\
    \midrule
    \multirow{8}{*}{\rotatebox[origin=c]{90}{CIFAR-10}} & \multirow{5}{*}{\rotatebox[origin=c]{90}{\linebreakcell{Wide\\Resnet-28-10}}} & \multirow{5}{*}{$96.5$\%} & Cutout & $=12$ & - & \\
        & & & Gaussian & - & $\leq0.1$ & \\
        & & & Patch Gaussian & $=25$ & $\leq1.0$ & \\
        \cmidrule(r){4-7}
        & & & Cutout* & $=22$ & - & \\
        & & & Gaussian* & - & $\leq1.0$ & \\
    \cmidrule(r){2-7}
    & \multirow{3}{*}{\rotatebox[origin=c]{90}{\linebreakcell{Shake\\112}}} & \multirow{3}{*}{$97.0$\%} & Cutout & $=7$ & - & \\
        & & & Gaussian & - & $\leq0.1$ & \\
        & & & Patch Gaussian & $=26$ & $\leq1.0$ & \\
    \midrule
    \multirow{13}{*}{\rotatebox[origin=c]{90}{ImageNet}} & \multirow{9}{*}{\rotatebox[origin=c]{90}{Resnet-50}} & \multirow{6}{*}{$76.0$\%} & Baseline & - & - & includes weight decay $=0.0001$\\
        & & & Cutout & $=60$ & - & \\
        & & & Gaussian & - & $\leq0.1$ & \\
        & & & Patch Gaussian & $\leq250$ & $\leq1.0$ & \\
        \cmidrule(r){4-7}
        & & & Cutout* & $=200$ & - & \\
        & & & Gaussian* & - & $\leq1.0$ & \\
        \cmidrule(r){4-7}
        & & & Larger Weight Decay & - & - & $0.001$\\
        & & & Dropblock & - & - & groups = $3$,$4$; keep prob = $0.9$\\
        & & & Label Smoothing & - & - & $0.1$\\
    \cmidrule(r){2-7}
    & \multirow{4}{*}{\rotatebox[origin=c]{90}{\linebreakcell{Resnet\\200}}} & \multirow{3}{*}{$78.5$\%} & Baseline & - & - & includes weight decay $=0.0001$\\
        & & & Cutout & $=30$ & - & \\
        & & & Gaussian & - & $\leq0.1$ & \\
        & & & Patch Gaussian & $\leq350$ & $\leq1.0$ & \\
    \bottomrule
  \end{tabular}
  }
\end{table}

\newpage
\begin{table}
  \caption{Full original corruption errors (Original CEs) for ImageNet models trained with different augmentation strategies.
  }
  \label{full-og-cc}
  \centering
  \begin{tabular}{cl|ccc|cccc}
    \toprule
        & & \multicolumn{3}{c|}{Noise} & \multicolumn{4}{c}{Blur}\\
        & Augmentation & Gaussian & Shot & Impulse & Defocus & Glass & Motion & Zoom\\
        \midrule
        \multirow{4}{*}{\rotatebox[origin=c]{90}{Resnet-50}} & Baseline & 0.705 & 0.722 & 0.716 & 0.815 & 0.915 & 0.810 & 0.817\\
        & Cutout & 0.720 & 0.727 & 0.720 & 0.798 & 0.923 & 0.821 & 0.813\\
        & Gaussian & 0.677 & 0.681 & 0.677 & 0.781 & \textbf{0.864} & 0.813 & 0.808\\
        & Patch Gaussian & \textbf{0.623} & \textbf{0.633} & \textbf{0.624} & \textbf{0.751} & 0.898 & \textbf{0.782} & \textbf{0.783}\\
        \midrule
        \multirow{4}{*}{\rotatebox[origin=c]{90}{Resnet-200}} & Baseline & 0.622 & 0.641 & 0.629 & 0.735 & 0.867 & 0.722 & 0.739 \\
        & Cutout & 0.594 & 0.619 & 0.600 & 0.714 & 0.870 & 0.713 & 0.737\\
        & Gaussian & 0.573 & 0.583 & 0.575 & 0.723 & 0.814 & 0.737 & 0.741\\
        & Patch Gaussian & \textbf{0.486} & \textbf{0.498} & \textbf{0.478} & \textbf{0.649} & \textbf{0.805} & \textbf{0.693} & \textbf{0.687}\\
        \bottomrule
  \end{tabular}
  \begin{tabular}{cl|cccc|cccc}
    \toprule
        & & \multicolumn{4}{c|}{Weather} & \multicolumn{4}{c}{Digital}\\
        & Augmentation & Snow & Frost & Fog & Bright & Contrast & Elastic & Pixel & JPEG \\
        \midrule
        \multirow{4}{*}{\rotatebox[origin=c]{90}{Resnet-50}} & Baseline & 0.827 & 0.756 & 0.589 & \textbf{0.582} & 0.748 & 0.753 & 0.799 & 0.747\\
        & Cutout & 0.839 & 0.764 & 0.599 & 0.586 & 0.747 & 0.752 & 0.803 & 0.752\\
        & Gaussian & 0.821 & \textbf{0.726} & 0.597 & 0.592 & 0.754 & \textbf{0.720} & 0.805 & 0.763\\
        & Patch Gaussian & \textbf{0.806} & 0.739 & \textbf{0.566} & 0.592 & \textbf{0.714} & 0.736 & \textbf{0.743} & \textbf{0.722}\\
        \midrule
        \multirow{4}{*}{\rotatebox[origin=c]{90}{Resnet-200}} & Baseline & 0.754 & 0.694 & 0.497 & 0.520 & 0.658 & 0.669 & 0.696 & 0.681\\
        & Cutout & 0.741 & 0.684 & 0.507 & 0.516 & 0.671 & 0.670 & 0.751 & 0.672\\
        & Gaussian & 0.731 & 0.653 & 0.525 & 0.514 & 0.699 & 0.641 & 0.693 & 0.660\\
        & Patch Gaussian & \textbf{0.697} & \textbf{0.633} & \textbf{0.476} & \textbf{0.506} & \textbf{0.627} & \textbf{0.625} & \textbf{0.613} & \textbf{0.593}\\
        \bottomrule
  \end{tabular}
\end{table}

\begin{table}[ht!]
  \caption{Full corruption errors (CEs) for ImageNet models trained with different augmentation strategies.
  }
  \label{full-cc}
  \centering
  \begin{tabular}{cl|ccc|cc}
    \toprule
        & & \multicolumn{3}{c|}{Noise} & \multicolumn{2}{c}{Blur}\\
        & Augmentation & Gaussian & Shot & Impulse & Defocus & Zoom\\
        \midrule
        \multirow{4}{*}{\rotatebox[origin=c]{90}{Resnet-50}} & Baseline & 1.000 & 1.000 & 1.000 & 1.000 & 1.000\\
        & Cutout & 1.015 & 1.013 & 1.008 & 0.979 & 1.000\\
        & Gaussian & 0.620 & 0.625 & 0.618 & 0.950 & 0.999\\
        & Patch Gaussian & \textbf{0.585} & \textbf{0.577} & \textbf{0.577} & \textbf{0.922} & \textbf{0.963}\\
        \midrule
        \multirow{4}{*}{\rotatebox[origin=c]{90}{Resnet-200}} & Baseline & 0.872 & 0.883 & 0.864 & 0.880 & 0.896\\
        & Cutout & 0.841 & 0.862 & 0.833 & 0.866 & 0.892\\
        & Gaussian & 0.533 & 0.538 & 0.538 & 0.855 & 0.910\\
        & Patch Gaussian & \textbf{0.490} & \textbf{0.488} & \textbf{0.490} & \textbf{0.767} & \textbf{0.820}\\
        \bottomrule
  \end{tabular}
  \begin{tabular}{cl|ccc|cccc}
    \toprule
        & & \multicolumn{3}{c|}{Weather} & \multicolumn{4}{c}{Digital}\\
        & Augmentation & Frost & Fog & Bright & Contrast & Elastic & Pixel & JPEG \\
        \midrule
        \multirow{4}{*}{\rotatebox[origin=c]{90}{Resnet-50}} & Baseline & 1.000 & 1.000 & 1.000 & 1.000 & 1.000 & 1.000 & 1.000\\
        & Cutout & 1.005 & 1.017 & 0.991 & 1.008 & 1.009 & 1.026 & 1.011\\
        & Gaussian & \textbf{0.919} & 1.073 & 1.019 & 1.051 & \textbf{0.967} & 0.974 & \textbf{0.966}\\
        & Patch Gaussian & 0.976 & \textbf{0.978} & \textbf{0.990} & \textbf{0.956} & 0.982 & \textbf{0.957} & 0.998\\
        \midrule
        \multirow{4}{*}{\rotatebox[origin=c]{90}{Resnet-200}} & Baseline & 0.912 & 0.862 & 0.888 & 0.861 & 0.882 & 0.848 & 0.922\\
        & Cutout & 0.915 & 0.868 & 0.877 & 0.877 & 0.871 & 0.875 & 0.911\\
        & Gaussian & 0.830 & 0.948 & 0.889 & 0.960 & 0.848 & 0.836 & 0.855\\
        & Patch Gaussian & \textbf{0.818} & \textbf{0.851} & \textbf{0.862} & \textbf{0.832} & \textbf{0.812} & \textbf{0.765} & \textbf{0.835}\\
        \bottomrule
  \end{tabular}
\end{table}

\begin{table}
  \caption{Comparison with SIN+IN \cite{geirhos2018imagenet}. By using Z=$74.6$\%, \aug{Patch Gaussian} can match SIN+IN's og mCE and test accuracy. Understandably, however, our gains are more concentrated in noise-based corruptions, whereas shape-biased models get gains in other corruptions.
  }
  \label{comparing-with-style}
  \centering
  \begin{tabular}{llccc}
    \toprule
        & Augmentation & Test Accuracy & og mCE & og mCE (-noise) \\
    \midrule
    \multirow{2}{*}{Resnet-50} & SIN+IN & 74.6\% & \textbf{0.693} & \textbf{0.699}\\
        & Patch Gaussian ($W\leq400,~\sigma\leq0.8$) & \textbf{75.6\%} & \textbf{0.693} & 0.718\\
        \bottomrule
  \end{tabular}
\end{table}

\begin{figure}[ht]
\begin{center}
  \centering
  \begin{tabular}{rccc}
    & \multicolumn{2}{c}{\linebreakcell{Wide Resnet\\CIFAR-10}} & \multicolumn{1}{c}{\linebreakcell{Resnet-50\\ImageNet}}                  \\
    \cmidrule(r){2-3} \cmidrule(r){4-4}
    & \linebreakcell{1st Layer\\Fourier\\Sensitivity} & \linebreakcell{Test Error\\Fourier\\Sensitivity} %
    & \linebreakcell{Test Error\\Fourier\\Sensitivity} \\
    \raisebox{2.0\normalbaselineskip}[0pt][0pt]{\rotatebox[origin=c]{90}{\aug{Cutout}}} & \includegraphics[width=0.12\textwidth]{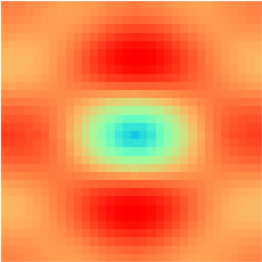} & \includegraphics[width=0.12\textwidth]{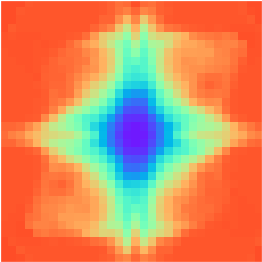} %
    & \includegraphics[width=0.12\textwidth]{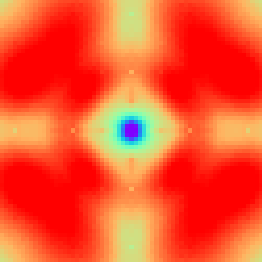}\\
    \raisebox{2.0\normalbaselineskip}[0pt][0pt]{\rotatebox[origin=c]{90}{\aug{Gaussian}}} & \includegraphics[width=0.12\textwidth]{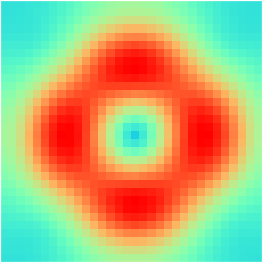} & \includegraphics[width=0.12\textwidth]{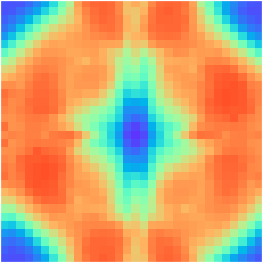}
    & \includegraphics[width=0.12\textwidth]{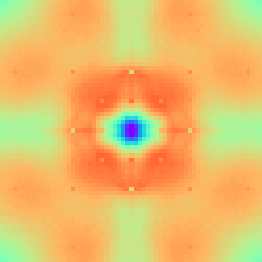}\\
    & \includegraphics[width=0.12\textwidth]{figures/fourier/colorbar.png} & \includegraphics[width=0.12\textwidth]{figures/fourier/colorbar.png} & \includegraphics[width=0.12\textwidth]{figures/fourier/colorbar_imagenet.png}\\
  \end{tabular}
\caption{Fourier analysis for \aug{Cutout} and \aug{Gaussian} models selected by the method in Section \ref{hyper-selection}. See Figure 
\ref{fourier} for details.}
\label{fourier-best-models}
\end{center}
\end{figure}

\begin{figure}[ht]
\begin{center}
  \centering
  \begin{tabular}{ccc}
    \aug{Baseline} & \aug{Cutout} & \aug{Gaussian}\\
    \cmidrule(r){1-1} \cmidrule(r){2-2} \cmidrule(r){3-3}
    \includegraphics[width=0.2\textwidth]{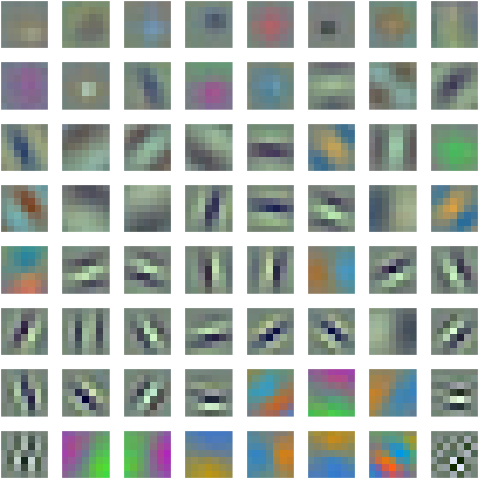} & \includegraphics[width=0.2\textwidth]{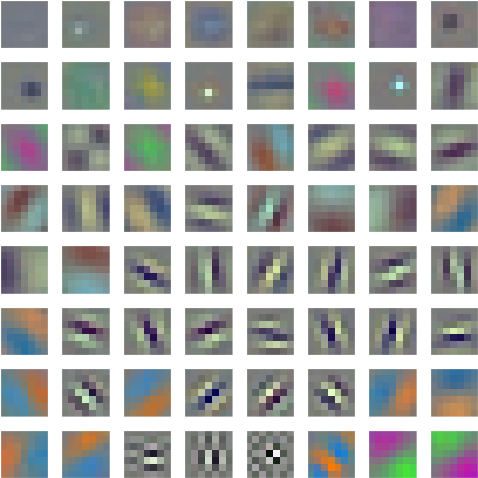} & \includegraphics[width=0.2\textwidth]{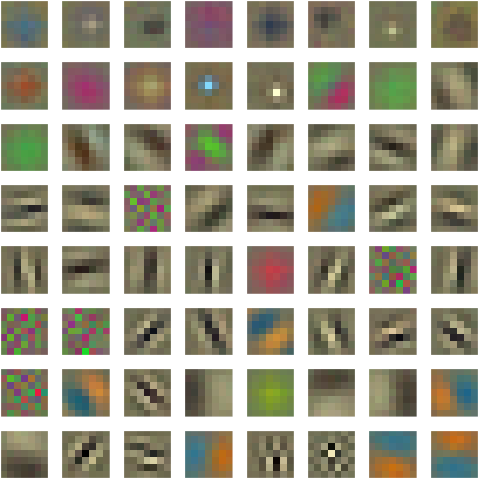}\\
    \\
    \aug{Patch Gaussian} & \aug{Cutout}* & \aug{Gaussian}*\\
    \cmidrule(r){1-1} \cmidrule(r){2-2} \cmidrule(r){3-3}
    \includegraphics[width=0.2\textwidth]{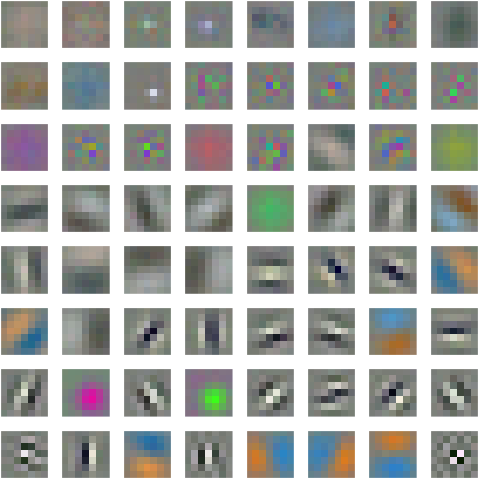} & \includegraphics[width=0.2\textwidth]{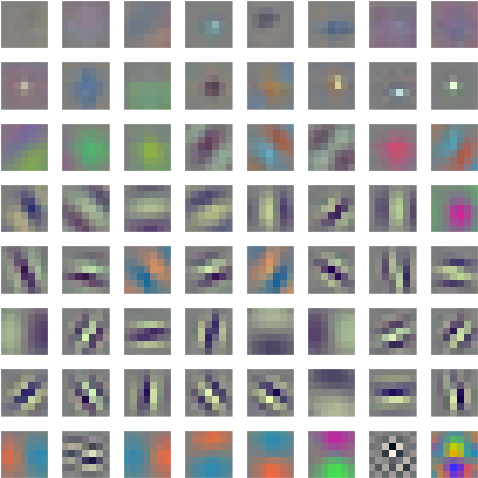} & \includegraphics[width=0.2\textwidth]{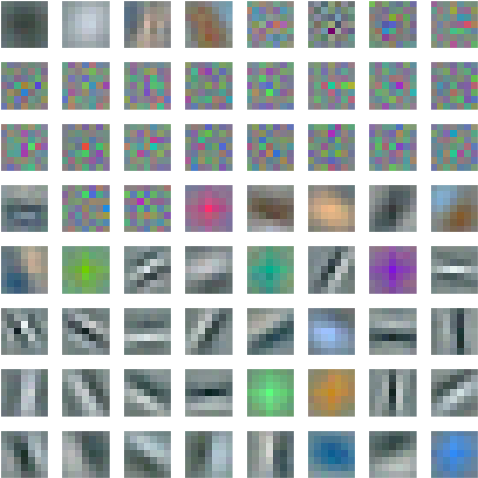}\\
  \end{tabular}
\caption{Complete filters for Resnet-50 models trained on ImageNet. *~Indicates augmentations with larger patch sizes and $\sigma$. See Figure \ref{fourier} for details. We again note the presence of filters of high fourier frequency in models trained with \aug{Cutout}* and \aug{Patch Gaussian}. We also note that \aug{Gaussian}* exhibits high variance filters. We posit these have not been trained and have little importance, given the low sensitivity of this model to high frequencies. Future work will investigate the importance of filters on sensitivity.
}
\label{full-filters}
\end{center}
\end{figure}

\begin{figure*}[h]
\begin{center}
\begin{minipage}[c]{0.7\textwidth}
\centerline{\includegraphics[width=\textwidth]{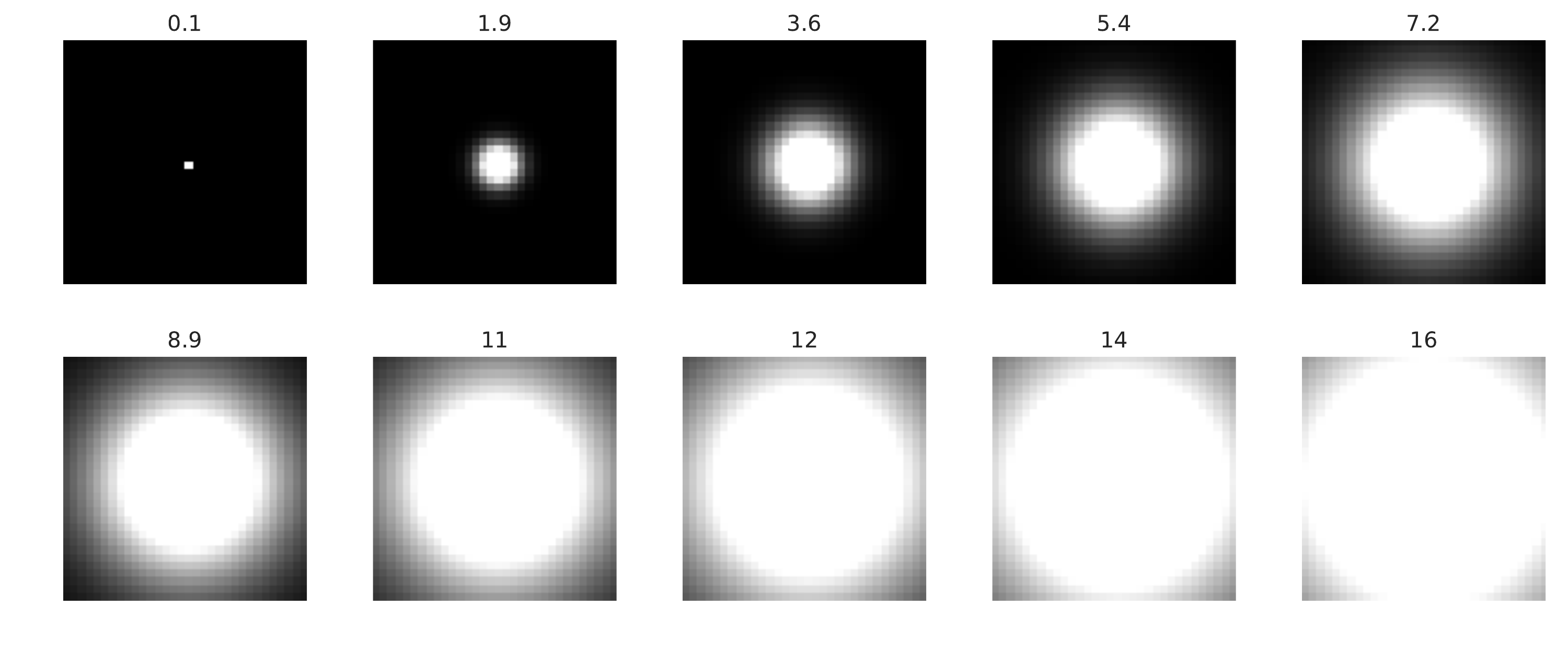}}
\end{minipage}%
\caption{Examples of high pass filters at various radii, in fourier space centered at the zero-frequency component, used in the high-pass experiment of Figure \ref{fourier}.
}
\label{highpass-deets}
\end{center}
\end{figure*}

\end{document}